\pdfoutput=1

\documentclass[11pt]{article}

\usepackage{acl}

\usepackage{times}
\usepackage{latexsym}

\usepackage[T1]{fontenc}

\usepackage[utf8]{inputenc}

\usepackage{microtype}

\usepackage{multirow}
\usepackage{booktabs,arydshln}
\usepackage{float}
\usepackage{graphicx}
\usepackage{caption,subcaption}
\usepackage{hyperref}
\usepackage{colortbl}
\usepackage{amsmath}
\usepackage{amsfonts}
\usepackage{pgfkeys}
\usepackage{tikz}
\usepackage{etoolbox}
\usepackage{pgf} 
\usepackage{xcolor} 
\usepackage{enumitem}
\usepackage{relsize}

\newcommand{\cmmnt}[1]{\ignorespaces}
\newcommand{\lang}[1]{\textsc{#1}}

\makeatletter
\def\adl@drawiv#1#2#3{%
        \hskip.5\tabcolsep
        \xleaders#3{#2.5\@tempdimb #1{1}#2.5\@tempdimb}%
                #2\z@ plus1fil minus1fil\relax
        \hskip.5\tabcolsep}
\newcommand{\cdashlinelr}[1]{%
  \noalign{\vskip\aboverulesep
           \global\let\@dashdrawstore\adl@draw
           \global\let\adl@draw\adl@drawiv}
  \cdashline{#1}
  \noalign{\global\let\adl@draw\@dashdrawstore
           \vskip\belowrulesep}}
\makeatother

\newcommand{\bm}[0]{\textit{Bi-LM}}
\newcommand{\mm}[0]{\textit{Mono-LM}}

\newcommand{\zs}[0]{\textit{BZ}}
\newcommand{\bs}[0]{\textit{BS}}
\newcommand{\mz}[0]{\textit{MZ}}

\newcommand{\supdiff}[0]{$\Delta_{(\textrm{BZ}-\textrm{BS})}$}

\newcommand{\monodiff}[0]{$\Delta_{(\textrm{MZ}-\textrm{BS})}$}

\newcommand{\realcorpus}[0]{$\mathbf{\mathcal{C}_{orig}}$}
\newcommand{\syncorpus}[0]{$\mathbf{\mathcal{C}_{deriv}}$}
\newcommand{\realdown}[1]{$\mathbf{\mathcal{D}_{\mathsmaller{orig}}^{#1}}$}
\newcommand{\syndown}[1]{$\mathbf{\mathcal{D}_{\mathsmaller{deriv}}^{#1}}$}

\newcommand{\trans}[0]{$\mathbf{\mathcal{T}}$}
\newcommand{\inv}[0]{$\mathbf{\mathcal{T}_{inv}}$}
\newcommand{\perm}[0]{$\mathbf{\mathcal{T}_{perm}}$}
\newcommand{\syntax}[0]{$\mathbf{\mathcal{T}_{syn}}$}
\newcommand{\script}[0]{$\mathbf{\mathcal{T}_{trans}}$}

\newcommand{\comp}[0]{$\mathbf{\circ}$}

\newcommand{\secref}[1]{(Section~\ref{#1})}
\newcommand{\symbolsecref}[1]{(\S~\ref{#1})}

\newcommand{\trm}[1]{\textrm{#1}}

\newcommand{\tablesep}[0]{\itemsep-0.2em}

\newcommand{\NA}{\cellcolor{low!5} ---}

\newcommand{\graytext}[1]{\textcolor{gray}{#1}}

\definecolor{approach}{HTML}{0070C0}
\newcommand*\circled[4]{\tikz[baseline=(char.base)]{
    \node[shape=circle, fill=#2, draw=#3, text=#4, inner sep=1.2pt, scale=0.9, font=\bfseries] (char) {#1};}}

\definecolor{high}{HTML}{9acd32}  
\definecolor{low}{HTML}{ff4500}  
\newcommand*{\minval}{-60}
\newcommand*{\maxval}{2}
\newcommand{\gradient}[1]{
    \ifdimcomp{#1pt}{>}{\maxval pt}{\cellcolor{high!30} #1}{
    \ifdimcomp{#1pt}{<}{\minval pt}{\cellcolor{low!80} #1}{
         \pgfmathparse{80-int(round(75*(#1/(\maxval-\minval))-(\minval*(75/(\maxval-\minval)))))}
        \xdef\tempa{\pgfmathresult}
        \cellcolor{low!\tempa} #1
    }}
 }
 
\newcommand*{\thresh}{0.2}
\newcommand*{\stepone}{2.0}
\newcommand*{\steptwo}{5.0}

\newcommand{\ApplyGradient}[1]{
        \ifdim #1 pt > \thresh pt
            \cellcolor{green!10}
            \ifdim #1 pt > \stepone pt
                \cellcolor{green!15}
            \fi
            \ifdim #1 pt > \steptwo pt
                \cellcolor{green!30}
            \fi
            #1
        \else
            \ifdim #1 pt < -\thresh pt
                \cellcolor{red!10}
                \ifdim #1 pt < -\stepone pt
                    \cellcolor{red!15}
                \fi
                \ifdim #1 pt < -\steptwo pt
                    \cellcolor{red!25}
                \fi
                #1                
            \else
                \cellcolor{yellow!15} #1
            \fi
        \fi
}

\title{When is BERT Multilingual? Isolating Crucial Ingredients for Cross-lingual Transfer}

\author{Ameet Deshpande \\ 
  Department of Computer Science \\
  Princeton University, USA \\
  {\small \texttt{asd@cs.princeton.edu}} \\
  \And
  Partha Talukdar \\
  Google Research \\
  India \\
  {\small \texttt{partha@google.com}} \\
  \And
  Karthik Narasimhan \\
  Department of Computer Science \\
  Princeton University, USA \\
   {\small \texttt{karthikn@cs.princeton.edu}} \\
}  

\begin{document}
\maketitle
\begin{abstract}
While recent work on multilingual language models has demonstrated their capacity for cross-lingual zero-shot transfer, there is a lack of consensus in the community as to what shared properties between languages enable transfer on downstream tasks.
Analyses involving pairs of natural languages are often inconclusive and contradictory since languages simultaneously differ in many linguistic aspects.
In this paper, we perform a large-scale empirical study to isolate the effects of various linguistic properties by measuring zero-shot transfer between four diverse natural languages and their counterparts constructed by modifying aspects such as the script, word order, and syntax.
Among other things, our experiments show that the absence of sub-word overlap significantly affects zero-shot transfer when languages differ in their word order, and there is a strong correlation between transfer performance and word embedding alignment between languages (e.g., $\rho_s=0.94$ on the task of NLI).
Our results call for focus in multilingual models on explicitly improving word embedding alignment between languages rather than relying on its implicit emergence.\footnote{Code is available at \url{https://github.com/princeton-nlp/MultilingualAnalysis}}


\end{abstract}

\section{Introduction}
\label{sec:introduction}

Multilingual language models like XLM \cite{DBLP:conf/acl/ConneauKGCWGGOZ20} and Multilingual-BERT~\cite{multilingualbert} are trained with masked-language modeling (MLM) objective on a combination of raw text from multiple languages.
Surprisingly, these models exhibit decent cross-lingual zero-shot transfer, where fine-tuning on a task in a source language translates to good performance for a different language (target). 

\paragraph{Requirements for zero-shot transfer}
Recent studies have provided inconsistent explanations for properties required for zero-shot transfer (hereon, transfer).
For example, while~\citet{wu2019beto} conclude that sub-word overlap is vital for transfer,~\citet{ketal} demonstrate that it is not crucial, although they consider only English as the source language.
While~\citet{pires2019multilingual} suggest that typological similarity (e.g., similar SVO order) is essential for transfer, other works~\cite{kakwani2020inlpsuite,DBLP:conf/acl/ConneauKGCWGGOZ20} successfully build multilingual models for dissimilar languages.


\paragraph{Need for systematic analysis}
A major cause of these discrepancies is a large number of varying properties (e.g., syntax, script, and vocabulary size) between languages, which make isolating crucial ingredients for transfer difficult.
Some studies alleviate this issue by creating synthetic languages which differ from natural ones only in specific linguistic properties like script~\cite{ketal,dufter2020identifying}.
However, their focus is only on English as a source language, and the scale of their experiments is small (in number of tasks or pre-training corpora size), thus limiting the scope of their findings to their settings alone.

\paragraph{Our approach}
We perform a systematic study of cross-lingual transfer on bilingual language models trained on a natural language and a systematically \textit{derived} counterpart.
We choose four diverse natural languages (English, French, Arabic, and Hindi) and create \textit{derived} variants using four different transformations on structural properties such as inverting or permuting word order, altering scripts, or varying syntax~\secref{sec:transformations:types}.
We train models on each of the resulting sixteen language pairs, and evaluate zero-shot transfer on four downstream tasks -- natural language inference (\lang{nli}), named-entity recognition (\lang{ner}), part-of-speech tagging (\lang{pos}), and question-answering (\lang{qa}).

\paragraph{Our experiments reveal the following:}
\begin{enumerate}[leftmargin=*]
    \tablesep
    \item Contrary to previous belief, the absence of sub-word overlap degrades transfer when languages differ in their word order (e.g., by more than $40$ F1 points on POS tagging,~\symbolsecref{sec:subword1}).
    \item There is a strong correlation between  token embedding alignment and zero-shot transfer across different tasks (e.g., $\rho_s=0.94,p<.005$ for XNLI, Fig~\ref{fig:xnli_correlation_annotated}).
    \item Using pre-training corpora from similar sources for different languages (e.g., Wikipedia) boosts transfer when compared to corpora from different sources (e.g., $17$ F1 points on NER, Fig~\ref{fig:diff_corpus}).
\end{enumerate}

To our knowledge, we are the first study to quantitatively show that zero-shot transfer between languages is strongly correlated with token embedding alignment ($\rho_s=0.94$ for \lang{nli}).
We also show that the current multilingual pre-training methods~\cite{doddapaneni2021primer} fall short of aligning embeddings even between simple natural and derived language pairs, leading to failure in zero-shot transfer.
Our results call for training objectives that explicitly improve alignment using either supervised (e.g., parallel corpora or bilingual dictionaries) or unsupervised data.

\section{Related work}
\label{sec:related}


\paragraph{Multilingual pre-training for Transformers}
The success of monolingual Transformer language models~\cite{devlin2019bert,radfordimproving} has driven studies that learn a multilingual language-model (LM) on several languages.
Multilingual-BERT (M-BERT)~\cite{multilingualbert} is a single neural network pre-trained using the masked language-modeling (MLM) objective on a corpus of text from $104$ languages.
XLM~\cite{DBLP:conf/nips/ConneauL19} introduced translation language-modeling, which performs MLM on pairs of parallel sentences, thus encouraging alignment between their representations.
These models exhibit surprising zero-shot cross-lingual transfer performance~\cite{DBLP:conf/nips/ConneauL19,ketal}, a setup where the model is fine-tuned on a source language and evaluated on a different target language.

\paragraph{Analysis of cross-lingual transfer}
While \citet{pires2019multilingual},~\citet{conneau2020emerging}, and~\citet{ketal} showed that transfer works even without a shared vocabulary between languages, \citet{wu2019beto} discovered a correlation between sub-word overlap and zero-shot performance.
\citet{conneau2020emerging} and \citet{artetxe2019cross} showed that shared parameters for languages with different scripts were crucial for transfer.
\citet{pires2019multilingual} and~\cite{wu2019beto} observed that transfer for NER and POS tagging works better between typologically similar languages. However, a study conducted by~\citet{lin2019choosing} showed that there is no simple rule of thumb to gauge when transfer works between languages.
\citet{hsu2019zero} observed that changing the syntax (SOV) order of the source to match that of the target does not improve performance.

\paragraph{Transfer between real and synthetic Languages}

\citet{ketal} create a synthetic language by changing English's script and find that transfer between it and Spanish works even without common sub-words.
However, they use only English as their source language, test only on two tasks, and use a single natural-synthetic language pair.
\citet{dufter2020identifying} study transfer between English and \textit{synthetic} English obtained by changing the script, word order, or model delimiters.
However, they use a small corpus ($228$K words) compared to current standards (we use $3$ orders more) and measure only embedding similarity and not zero-shot transfer.
A contemporary work~\cite{wu2022oolong} uses synthetic transformations to modify the GLUE dataset~\cite{wang2018glue} and analyze properties required for good zero-shot transfer, but they perform their experiments only on English and do not perform token embedding alignment analysis.
We show that the latter is crucial for good transfer.
\section{Approach}
\label{sec:approach}


\begin{table*}[t]
\centering
\begin{tabular}{@{}ll@{\hskip 3em}l@{}}
\toprule
\textbf{Transformation} & \textbf{Instance ($s$)}                                      & \textbf{Transformed instance $\left ( \textrm{\trans}(s) \right )$}                        \\ \midrule
\textbf{\textit{Inversion}} (\inv) & Welcome to NAACL at Seattle                               & Seattle at NAACL to Welcome                             \\

\textbf{\textit{Permutation}} (\perm) & This is a conference                   & a This conference is                 \\

\textbf{\textit{Transliteration}} (\script) & I am Sam . I am                                        & $\clubsuit_\textrm{\graytext{(I)}}\; \heartsuit_\textrm{\graytext{(am)}}\; \diamondsuit_\textrm{\graytext{(Sam)}}\; \spadesuit_\textrm{\graytext{(.)}}\; \clubsuit_\textrm{\graytext{(I)}}\; \heartsuit_\textrm{\graytext{(am)}}\; $                                         \\ 

\multirow{2}{*}{\textbf{\textit{Syntax}} (\syntax)} & Sara {\small \graytext{(S)}} ate {\small \graytext{(V)}} apples {\small \graytext{(O)}} & Sara {\small \graytext{(S)}} apples {\small \graytext{(O)}} ate {\small \graytext{(V)}} \\

 & Une table {\small \graytext{(N)}} ronde {\small \graytext{(A)}}  & Une ronde {\small \graytext{(A)}} table {\small \graytext{(N)}} \\

\bottomrule
\end{tabular}%
\caption{
Examples of our transformations applied to different sentences (without sub-word tokenization).
\textit{Inversion} inverts the tokens, \textit{Permutation} samples a random reordering, and \textit{Transliteration} changes the script.
We use symbols ($\clubsuit$) to denote words in the new script and mention the corresponding original word in brackets.
\textit{Syntax} stochastically modifies the syntactic structure.
In the first example for \textit{Syntax}, the sentence in Subject-Verb-Object (SVO) order gets transformed to SOV order, and in the second, the sentence in Noun-Adjective (NA) order gets transformed to the AN order.
The examples are high probability re-orderings and other ones might be sampled too.
}
\label{table:trans_examples}
\end{table*}

We first provide some background on bilingual language models (Section~\ref{sec:background}), followed by descriptions of our transformations (Section~\ref{sec:transformations}), and our training and evaluation setup (Section~\ref{sec:models}).

\subsection{Background}
\label{sec:background}

\paragraph{Bilingual pre-training}
\label{sec:bilm_setup:pretraining}

The standard setup~\cite{DBLP:conf/nips/ConneauL19} trains a bilingual language model (\bm{}) on raw text corpora from two languages simultaneously.
\bm{} uses the masked language-modeling loss ($\mathcal{L}_{\textrm{MLM}}$) on the corpora from the two languages ($\mathcal{C}_1, \mathcal{C}_2$) separately with no explicit cross-lingual signal:

\vspace{-1.5em}
\begin{align*}
    \mathcal{L}^{\theta}_{\textrm{Bi-LM}}( \mathcal{C}_{\mathbf{1}} + \mathcal{C}_{\mathbf{2}} ) = \mathcal{L}^{\theta}_{\textrm{MLM}}( \mathcal{C}_{\mathbf{1}} ) + \mathcal{L}^{\theta}_{\textrm{MLM}}\left ( \mathcal{C}_{\mathbf{2}} \right )
\end{align*}
A shared byte pair encoding tokenizer~\cite{sennrich2015neural} is trained on $\mathcal{C}_1 + \mathcal{C}_2$.
A single batch contains instances from both languages, but each instance belongs to a single language.

\paragraph{Zero-shot transfer evaluation}
\label{sec:bilm_setup:zero_shot}

Consider a bilingual model (\bm) pre-trained on two languages, \textit{source} and \textit{target}.
Zero-shot transfer involves fine-tuning \bm{} on downstream task data from \textit{source} and evaluating on test data from \textit{target}.
This is considered zero-shot because \bm{} is not fine-tuned on any data belonging to \textit{target}.
\subsection{Generating language variants with systematic transformations}
\label{sec:transformations}

\begin{figure*}[ht]
\begin{subfigure}{.47\linewidth}
  \centering
  \includegraphics[width=\textwidth]{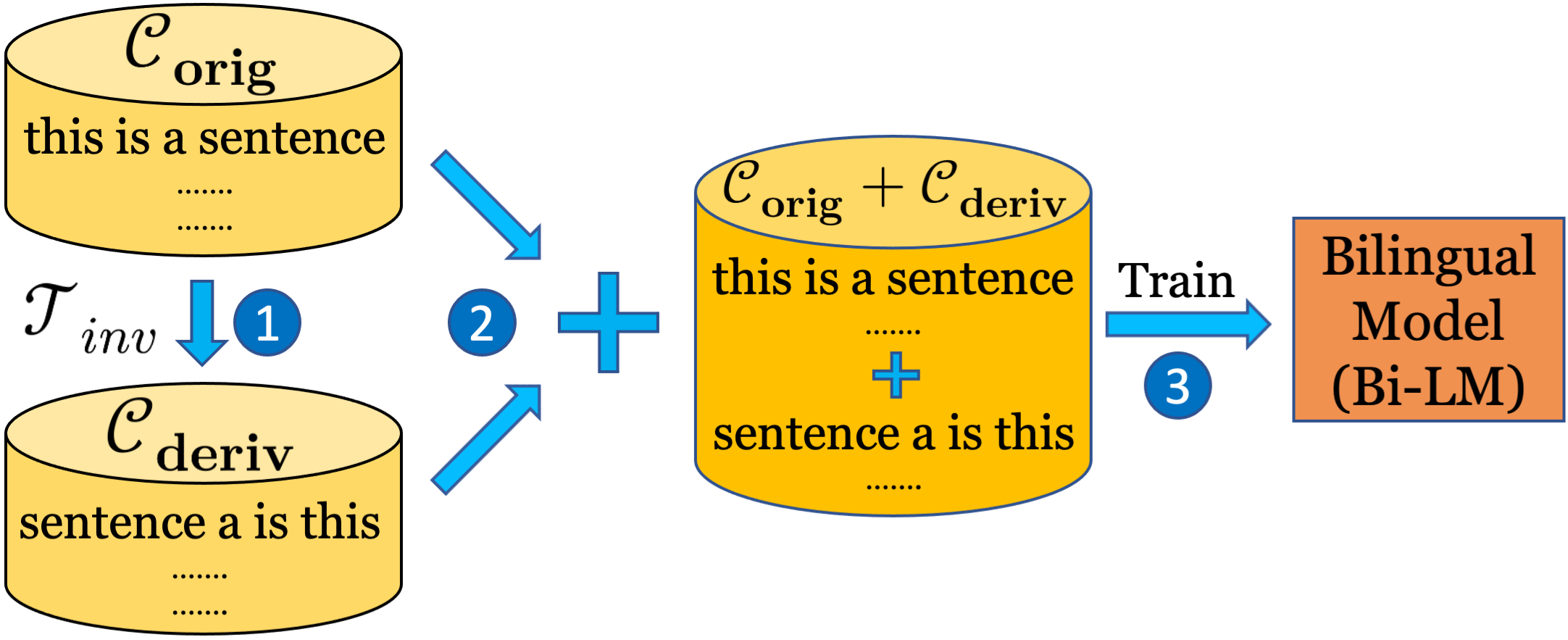}
  \caption{Pre-training}
  \label{fig:pretraining}
\end{subfigure}\hfill
\begin{subfigure}{.47\linewidth}
  \centering
  \includegraphics[width=\textwidth]{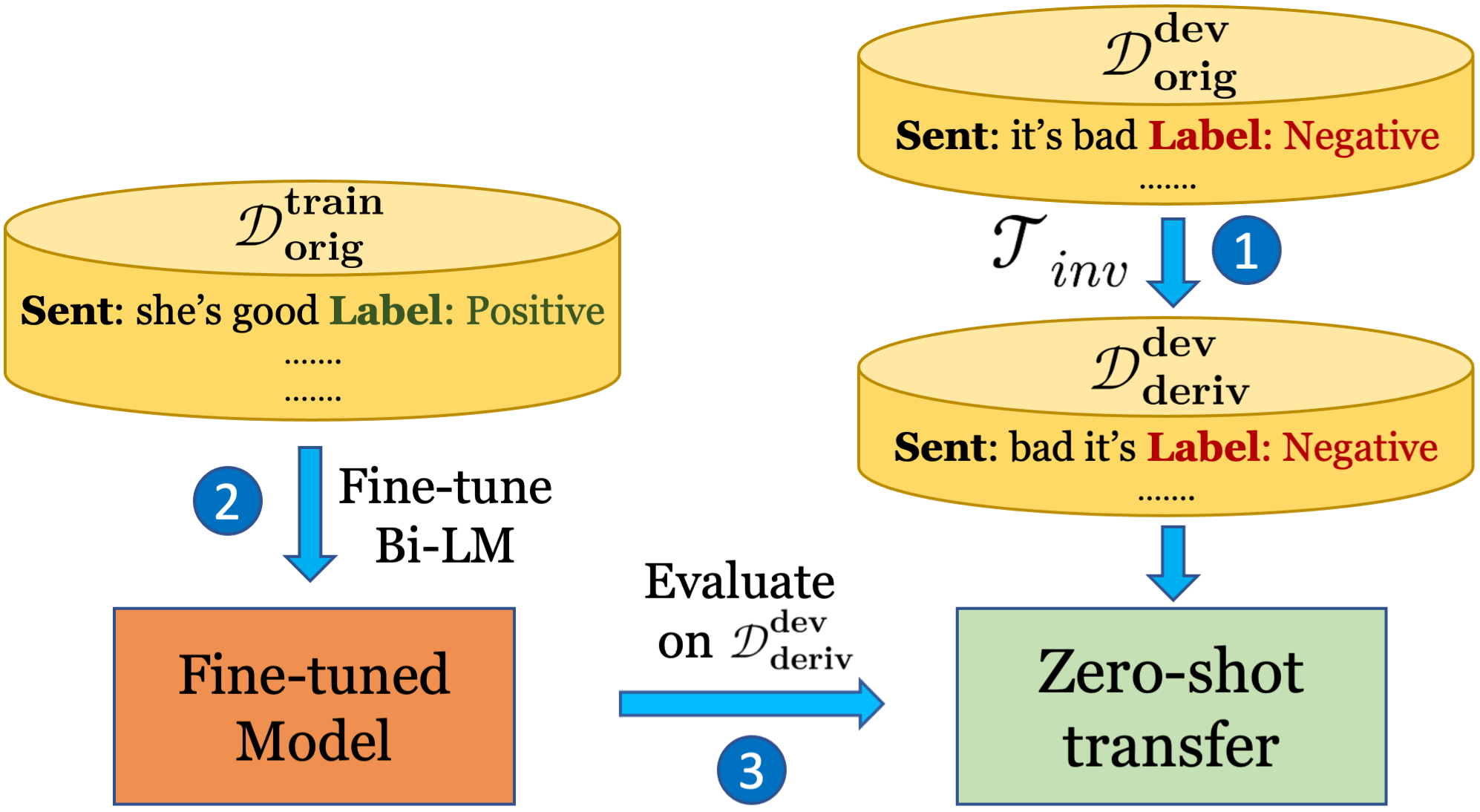}
  \caption{Fine-tuning}
  \label{fig:finetuning}
\end{subfigure}
\caption{
\textbf{(a)} During pre-training, we \circled{1}{approach}{approach}{white} obtain the \textit{derived} language corpus (\syncorpus) by \textit{transforming} the \textit{original} language corpus (\realcorpus).
\circled{2}{approach}{approach}{white} The two corpora are combined and,
\circled{3}{approach}{approach}{white} a bilingual model (\bm) is learned using the MLM objective.
\textbf{(b)} During fine-tuning, we \circled{1}{approach}{approach}{white} obtain the derived dev dataset (\syndown{dev}) by transforming the original dev dataset (\realdown{dev}).
\circled{2}{approach}{approach}{white} \bm{} is fine-tuned on the original train dataset (\realdown{train}), and
\circled{3}{approach}{approach}{white} evaluated on \syndown{dev}, which is the standard zero-shot cross lingual setup.
}
\label{fig:approach}
\end{figure*}

Natural languages typically differ in several ways, like the script, word order, and syntax.
To isolate the affect of these properties on zero-shot transfer, we obtain \textit{derived} language corpora (hereon, \textit{derived} corpora) from \textit{original} (natural) language corpora by performing sentence level transformations (\trans) which change particular properties.
For example, an ``\textit{inversion}'' transformation could be used to invert each sentence in the corpus (\textit{Welcome$_1$ to$_2$ NAACL$_3$} $\Rightarrow$ \textit{NAACL$_3$ to$_2$ Welcome$_1$}).
Since the transformation (\trans) is applied on each sentence of the \textit{original} corpus, the size of the \textit{original} and the \textit{derived} corpus is the same.
In the following sections, we will use the following notation:
\begin{align*}
    \begin{split}
        \textrm{\realcorpus{}} &\equiv \textrm{Original corpus} \\
         &= \{ s_i \; \vert \; i=1:N, s_i=\textrm{sentence} \} \\  
        \textrm{\trans{}} &\equiv \textrm{Sentence-level transformation} \\
        \textrm{\syncorpus{}}  &\equiv \textrm{Derived corpus} \\
         &= \{ \textrm{\trans(sent)} \; \vert \; \forall \: \textrm{sent} \in \textrm{\realcorpus{}} \}
    \end{split}
\end{align*}\vspace{-1em}

\paragraph{Types of transformations}
\label{sec:transformations:types}

We consider four transformations which modify different aspects of sentences (examples in Table~\ref{table:trans_examples}):

\begin{enumerate}
    \tablesep
    \item \textbf{Inversion} (\inv): Invert the order of \textit{tokens} in the sentence, like in~\citet{dufter2020identifying}. The first token becomes the last, and vice versa.
    \item \textbf{Permutation} (\perm): Permute the order of tokens in a sentence uniformly at random.
    For a sentence of $n$ tokens, we sample a random ordering with probability $\frac{1}{n!}$.
    \item \textbf{Transliteration} (\script): Change the script of all tokens other than the special tokens (like \texttt{[CLS]}).
    This creates a \textit{derived} vocabulary ($\mathcal{V}_{deriv}$) with a one-to-one correspondence with the original vocabulary ($\mathcal{V}_{orig}$).
    \item \textbf{Syntax} (\syntax): Modify a sentence to match the syntactic properties of a different natural language by re-ordering the dependents of nouns and verbs in the dependency parse.
    These transformations are stochastic because of the errors in parsing and sampling over possible re-orderings~\cite{wang2016galactic}.

\end{enumerate}

These transformations allow us to systematically evaluate the effect of corresponding properties on zero-shot transfer.
We also consider composed transformations~($\S$\ref{sec:subword2}) which consecutively apply two transformations.
We note that while real languages typically differ in more than one or two properties considered in our transformations, our methodology remains useful in isolating crucial properties that enable good transfer and can be extended to more transformations.


\paragraph{Transformations for downstream tasks}
\label{sec:trans_downstream}

We obtain the downstream corpus in the \textit{derived} language (\syndown{}) by applying the same transformation (\trans) used during pre-training on the \textit{original} downstream corpus (\realdown{}). 
Unlike pre-training corpora which contain raw sentences, instances in downstream tasks contain one or more sentences with annotated labels.
For text classification tasks like NLI, we apply the transformation on each sentence in every dataset instance.
For token classification tasks (e.g., NER, POS), any transformation which changes the order of the tokens also changes the order of the labels.
We present the mathematical specification in Appendix~\ref{appendix:MathematicalDownstream}.
\subsection{Model Training and Evaluation}
\label{sec:models}

\begin{table}
\centering
\resizebox{1\columnwidth}{!}{
\begin{tabular}{cccc}
\toprule

\multirow{2}{*}{\textbf{Evaluation}} & \multicolumn{3}{c}{\textbf{Corpus source}} \\ \cmidrule{2-4}
 & \textbf{Pre-train} & \textbf{Fine-tune (train)} & \textbf{Fine-tune (dev)} \\ \midrule
\textbf{\colorbox{red!15}{BZ}} & \realcorpus{} $+$ \syncorpus{} & \realdown{} & \syndown{} \\ \specialrule{.0em}{.2em}{.2em}

\textbf{\colorbox{teal!15}{BS}} & \realcorpus{} $+$ \syncorpus{} & \textcolor{blue}{\syndown{}} & \syndown{} \\ \specialrule{.0em}{.2em}{.2em}

\textbf{\colorbox{green!15}{MZ}} & \textcolor{blue}{\realcorpus{}} & \realdown{} & \syndown{} \\ \midrule

\multicolumn{4}{c}{$\trm{\supdiff} = \left ( \textrm{\textbf{\colorbox{red!15}{BZ}}} - \textrm{\textbf{\colorbox{teal!15}{BS}}} \right )$} \\
\multicolumn{4}{c}{$\trm{\monodiff} = \left ( \textrm{\textbf{\colorbox{green!15}{MZ}}} - \textrm{\textbf{\colorbox{teal!15}{BS}}} \right )$} \\

\bottomrule

\end{tabular}
}
\caption{
Summary of evaluation metrics defined in \S~\ref{subsection:Evaluation}.
$\mathbf{\mathcal{C}}$ and $\mathbf{\mathcal{D}}$ denote the pre-training and downstream corpus respectively, and their subscript indicates their source (\textit{original} or \textit{derived}).
\textbf{BZ} and \textbf{MZ} represent bilingual and monolingual zero-shot transfer scores, and \textbf{BS} is the supervised learning baseline on \textit{derived}.
The differences in the setting of \textbf{BZ} and other scores are typeset in blue.
We use \textbf{\supdiff{}} and \textbf{\monodiff{}} (defined in the last two rows) throughout our paper.
}
\label{table:EvalExplanation}
\end{table}

We now describe our pre-training and zero-shot transfer evaluation setup.
Figure~\ref{fig:approach} provides an overview of pre-training and fine-tuning, and Table~\ref{table:EvalExplanation} summarizes the evaluation metrics we use.


\paragraph{Pre-training}
\label{sec:models:types}

Let \realcorpus{} and \syncorpus{} be the \textit{original} and \textit{derived} language pre-training corpora.
We train two models for each \textit{original}-\textit{derived} pair:
\begin{enumerate}
    \tablesep
    \item \textbf{Bilingual Model (\bm)}: A bilingual model pre-trained on the combined corpus (\realcorpus $+$ \syncorpus)
    (Figure~\ref{fig:pretraining}).
    \item \textbf{Monolingual Model (\mm)}: A monolingual model trained only on \realcorpus{} for the same number of steps as \bm{}'s.
    \mm{} is used as a baseline to measure zero-shot transfer of a model not pre-trained on \textit{derived}.
\end{enumerate}


\paragraph{Evaluation}
\label{subsection:Evaluation}

Let \realdown{train} and \realdown{dev} be the \textit{original} language training and development sets for a downstream task,
and \syndown{train} and \syndown{dev} be the corresponding \textit{derived} language datasets.
For evaluation, we first fine-tune the pre-trained models on a downstream training set and evaluate the resulting model on a development set (Figure~\ref{fig:finetuning}).
Since our goal is to investigate the extent of zero-shot transfer, we require appropriate lower and upper bounds to make informed conclusions.
To this end, we compute three metrics, all on the same development set (summarized in Table~\ref{table:EvalExplanation}):
\begin{itemize}
    \tablesep
    \item \textbf{Bilingual zero-shot transfer (\zs)}: This is the standard zero-shot transfer score~\cite{DBLP:conf/nips/ConneauL19} which measures how well a bilingual model fine-tuned on \realdown{train} zero-shot transfers to the other language (\syndown{dev}).

    \item \textbf{Bilingual supervised synthetic (\bs)}: 
    This is the supervised learning performance on the \textit{derived} language obtained by fine-tuning \bm{} on \syndown{train} and evaluating it on \syndown{dev}.

    \item \textbf{Monolingual zero-shot transfer (\mz)}:
    This measures the zero-shot performance of the baseline \mm{}, which is not pre-trained on the \textit{derived} language, by fine-tuning \mm{} on \realdown{train} and evaluating it on \syndown{dev}.

\end{itemize}
\bs{} uses fine-tuning train data from the \textit{derived} language and serves as an upper-bound on \zs{} and \mz{} which don't use it.
\mz{} doesn't pre-train on the \textit{derived} language and serves as a lower-bound on \zs{} which does pre-train on it. 
For easier comparison of \zs{} and \mz{} with \bs{} (upper-bound), we report the following score differences (Table~\ref{table:EvalExplanation}), which are both negative in our experiments.

\vspace{-1.5em}
\begin{align}
        \trm{\supdiff} &= \left ( \textrm{BZ} - \textrm{BS} \right ) \\
        \trm{\monodiff} &= \left ( \textrm{MZ} - \textrm{BS} \right )
\end{align}

\zs{} alone cannot capture the quality of the zero-shot transfer.
A large and negative \supdiff{} implies that bilingual zero-shot transfer is much worse than supervised fine-tuning on \textit{derived}.
Concurrently, \supdiff{} $\approx$ \monodiff{} implies that \bm{} transfers as poorly as \mm{}.
\textbf{Thus, good zero-shot transfer is characterized by \supdiff{} $\approx 0$ and \supdiff{} $\gg$ \monodiff{}.}
\begin{table*}[t]
\centering
\resizebox{\linewidth}{!}{
\begin{tabular}{ccccccccccccc}\toprule
\multirow{2}{*}{\textbf{Task}}  & \multicolumn{3}{c}{\textbf{Inversion (\inv)}} & \multicolumn{3}{c}{\textbf{Permutation (\perm)}} & \multicolumn{3}{c}{\textbf{Syntax (\syntax)}} & \multicolumn{3}{c}{\textbf{Transliteration (\script)}}\\
\cmidrule(lr){2-4} \cmidrule(lr){5-7} \cmidrule(lr){8-10} \cmidrule(lr){11-13}


& \textbf{\small{\supdiff}} & \textbf{\small{\monodiff}} & \graytext{\textbf{\zs}} & \textbf{\small{\supdiff}} & \textbf{\small{\monodiff}} & \graytext{\textbf{\zs}} & \textbf{\small{\supdiff}} & \textbf{\small{\monodiff}} & \graytext{\textbf{\zs}} & \textbf{\small{\supdiff}} & \textbf{\small{\monodiff}} & \graytext{\textbf{\zs}}\\ \midrule


 \textbf{XNLI} &  \gradient{-10.2} & \gradient{-13.0} & \graytext{58.4}  &  \gradient{-3.6} & \gradient{-8.6} & \graytext{62.6}  &  \gradient{-0.9}$^\star$ & \gradient{-1.1} & \graytext{67.8}  & \gradient{-1.0}$^\star$ & \gradient{-36.7} & \graytext{69.3} \\ \specialrule{.0em}{.5em}{.5em} 
 
\textbf{NER} &  \gradient{-49.1} & \gradient{-46.7} & \graytext{37.9}  & \gradient{-26.3} & \gradient{-35.4} & \graytext{47.3}  & \gradient{-14.6} & \gradient{-16.6} & \graytext{62.9}  & \gradient{-1.9}$^\star$ & \gradient{-82.6} & \graytext{83.7} \\ \specialrule{.0em}{.5em}{.5em} 
 
\textbf{POS}  & \gradient{-30.2} & \gradient{-36.2} & \graytext{64.2}  & \gradient{-11.2} & \gradient{-25.2} & \graytext{73.6} & \gradient{-4.4} & \gradient{-7.6} & \graytext{89.4}  & \gradient{-0.4}$^\star$ & \gradient{-95.0} & \graytext{95.4} \\ \specialrule{.0em}{.5em}{.5em} 

\textbf{XQuAD}\footnotemark[4] & \gradient{-32.8} & \gradient{-31.0} & \graytext{22.8} & \NA\footnotemark[4]  & \NA & \graytext{---} & \NA\footnotemark[4] & \NA & \graytext{---} & \gradient{0.0}$^\star$ & \gradient{-55.9} & \graytext{61.2}\\

\bottomrule
\end{tabular}
}
\caption{
\label{table:main_table}
\textbf{(1) Evaluation: }
We report \textbf{\supdiff{}} and \textbf{\monodiff{}} (\S~\ref{subsection:Evaluation} and Table~\ref{table:EvalExplanation}) for transformations on different tasks, averaged over four languages (\lang{en}, \lang{fr}, \lang{hi}, \lang{ar}).
We report the breakdown for different languages in Appendix~\ref{appendix:sec:joint_table}.
\zs{}, the bilingual zero-shot performance, is reported for reference.
\textbf{(2) Interpreting scores: }
Smaller (more negative) \supdiff{} implies worse bilingual zero-shot transfer, whereas \supdiff{}$\approx 0$ implies strong transfer.
\supdiff{} $\gg$ \monodiff{} implies that bilingual pre-training is extremely useful.
Scores are highlighted based on their value (lower scores have a higher intensity of {\setlength{\fboxsep}{2pt}\colorbox{low!40}{red}}).
Cases with strong zero-shot transfer (\supdiff{}$\approx 0$) are marked with an asterisk.
\textbf{(3) Trends: }
\script{} exhibits strong transfer on all tasks and languages (high \supdiff{} scores), and bilingual pre-training is extremely useful (\supdiff{} $\gg$ \monodiff{}), implying that zero-shot transfer is possible between languages with different scripts but the same word order.
\inv{} and \perm{} suffer on all tasks (small \supdiff{} scores) whereas \syntax{} suffers significantly lesser, which provides evidence that local changes to the word order made by \textit{Syntax} (\syntax{}) hurts zero-shot transfer significantly lesser than global changes made by \textit{Inversion} (\inv{}) and \textit{Permutation} (\perm{}).
}
\end{table*}

\subsection{Experimental Setup}
\label{sec:experiments}


\begin{table}
\centering
\resizebox{1\columnwidth}{!}{
\begin{tabular}{c c c}
\toprule

\textbf{Dataset} & \textbf{Task} & \textbf{Metric} \\ \midrule
XNLI~\cite{conneau2018xnli} & NLI & Accuracy \\
Wikiann~\cite{pan2017cross} & NER & F1 \\
UD v2.5~\cite{11234/1-2837} & POS & F1 \\
XQuAD~\cite{artetxecross} & QA & F1 \\

\bottomrule
\end{tabular}
}
\caption{
XTREME benchmark datasets used for zero-shot transfer evaluation. NLI=Natural Language Inference, NER=Named-entity recognition, POS=Part-of-speech tagging, QA=Question-Answering.
}
\label{table:downstream_datasets}
\end{table}



\paragraph{Languages} We choose four diverse natural languages: English (Indo-European, Germanic), French (Indo-European, Romance), Hindi (Indo-European, Indo-Iranian), and Arabic (Afro-Asiatic, Semitic), which are represented in the multilingual XTREME benchmark~\cite{hu2020xtreme}.
For each language, we consider four transformations (Section~\ref{sec:transformations:types}) to create \textit{derived} counterparts, giving us $16$ different original-derived pairs in total.
For the \textit{Syntax} transformation, we use~\citet{qi2020stanza} for parsing.
We modify the syntax of \lang{fr}, \lang{hi}, and \lang{ar} to that of \lang{en}, and the syntax of \lang{en} to that of \lang{fr}.

\paragraph{Datasets} For the pre-training corpus (\realcorpus), we use a $500$MB (uncompressed) subset of Wikipedia ($\approx$ 100M tokens) for each language.
This matches the size of WikiText-103~\cite{merity2016pointer}, a standard language-modeling dataset.
For downstream evaluation, we choose four tasks from the XTREME benchmark~\cite{hu2020xtreme}.
Table~\ref{table:downstream_datasets} lists all the datasets and their evaluation metrics.

\paragraph{Implementation Details}
\label{subsection:TrainingDetails}

We use a variant of  RoBERTa~\cite{liu2019roberta} which has $8$ layers, $8$ heads, and a hidden dimensionality of $512$.
We train each model on $500$K steps, a batch size of $128$, and a learning rate of $1e\textrm{-}4$ with a linear warmup of $10$K steps.
We use an \textit{original} language vocabulary size of $40000$ for all the models and train on $8$ Cloud TPU v3 cores for $32$-$48$ hours.
For fine-tuning, we use standard hyperparameters (Appendix~\ref{appendix:xtreme}) from the XTREME benchmark and report our scores on the development sets.


\begin{figure*}[ht]

\centering
\begin{subfigure}[b]{0.32\textwidth}
\centering
\includegraphics[width=\textwidth]{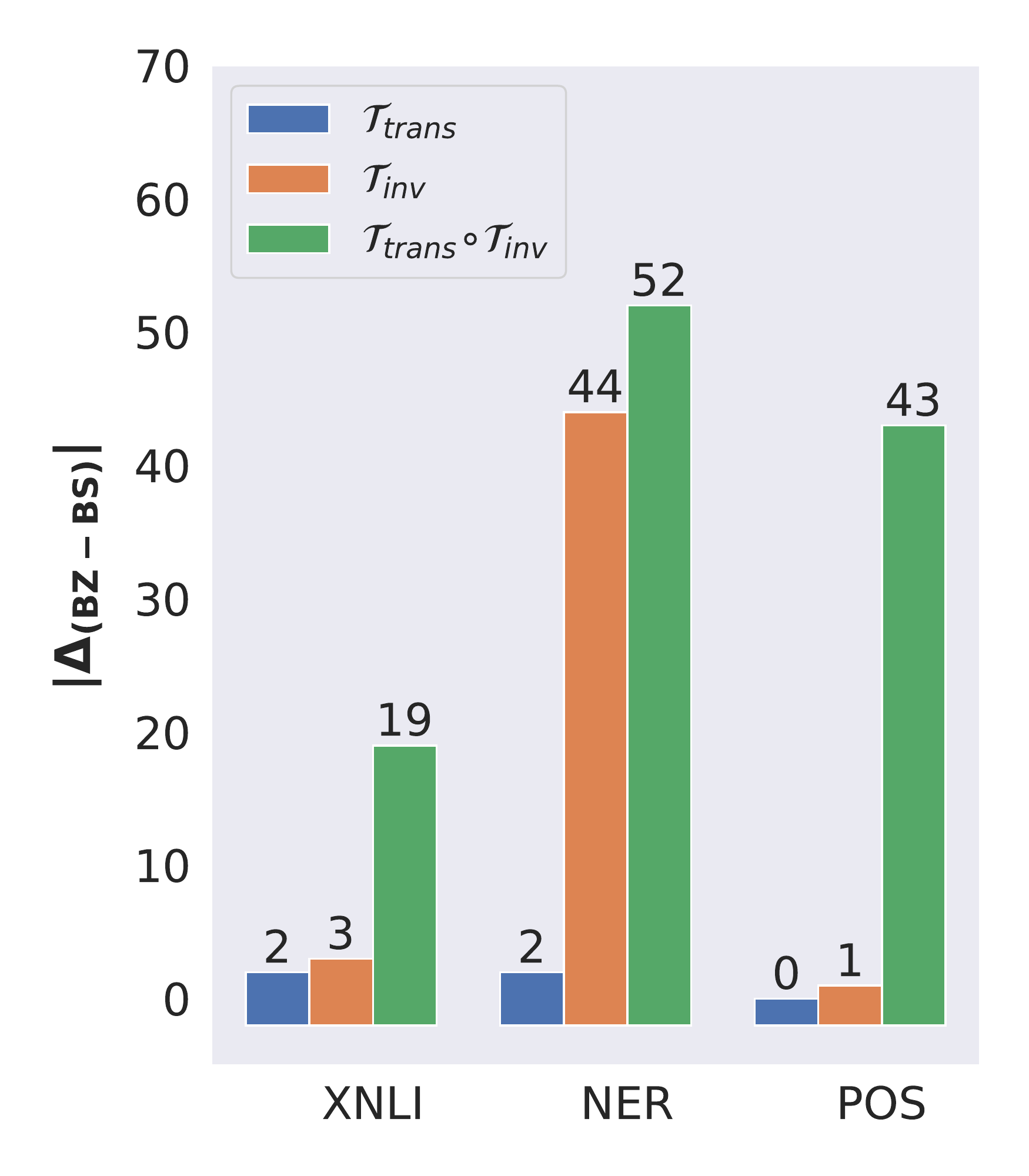}
\end{subfigure}
\begin{subfigure}[b]{0.32\textwidth}
\centering
\includegraphics[width=\textwidth]{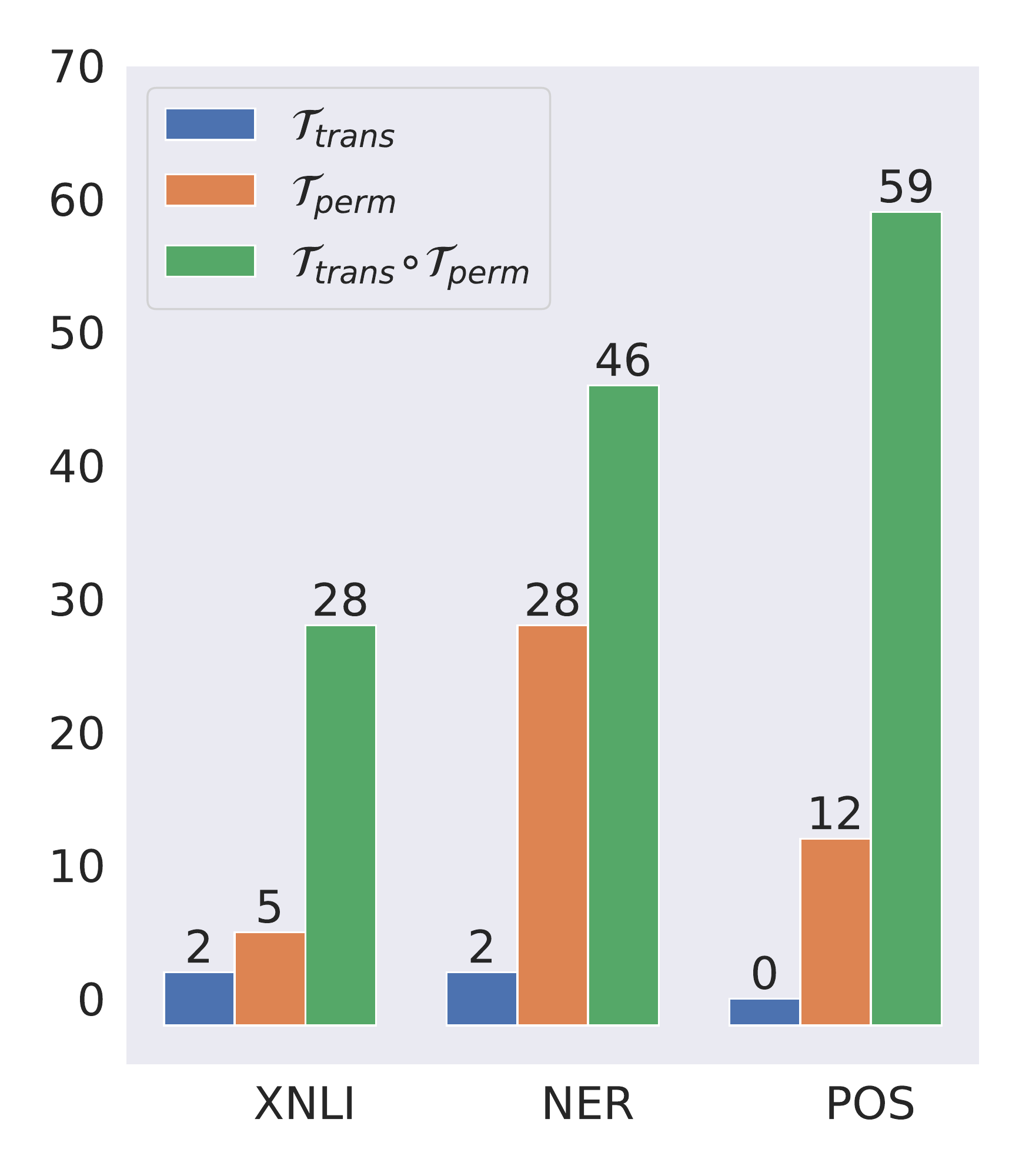}
\end{subfigure}
\begin{subfigure}[b]{0.32\textwidth}
\centering
\includegraphics[width=\textwidth]{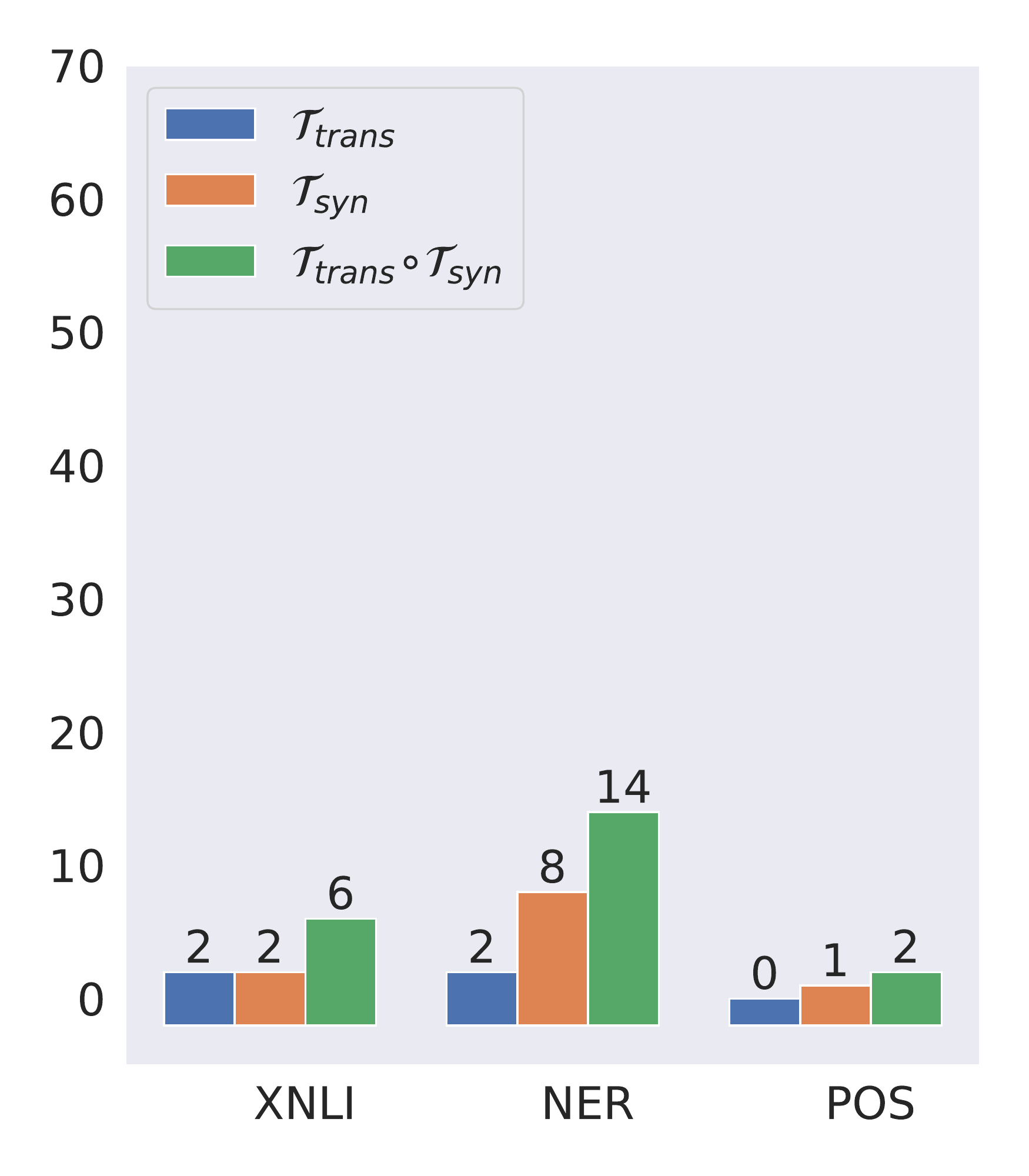}
\end{subfigure}

\caption{
$| \textrm{\supdiff{}} |$ for composed transformations~\symbolsecref{sec:subword2} applied on \lang{en} as the \textit{original} language.
Larger scores imply worse zero-shot transfer.
\script{} = \textit{Transliteration}, \inv{} = \textit{Inversion}, \perm{} = \textit{Permutation}, and \syntax{} = \textit{Syntax}.
Sub-word overlap between the \textit{original} and \textit{derived} language is $0\%$ when composed transformations are used (e.g. \script{} $\circ$ \inv{}) and $100\%$ when the second constituent is used (here, \inv{}).
We observe that the composed transformations (green bars) do significantly worse than their constituents (blue and orange), showing that
\script{} $\circ$ \inv{} is worse than \inv{} by over $16$ points on XNLI and $42$ points on POS, with similar trends for \script{} $\circ$ \perm{}.
\script{} $\circ$ \syntax{} doesn't suffer as much, but its performance degradation when compared to \textit{Syntax} is still large (ranges between $1$ point on POS to $6$ points on NER).
\textbf{absence of sub-word overlap significantly hurts performance when languages differ in their word orders.}
}
\label{fig:composed}

\end{figure*}

\begin{figure}[ht]
\centering
\includegraphics[width=\linewidth]{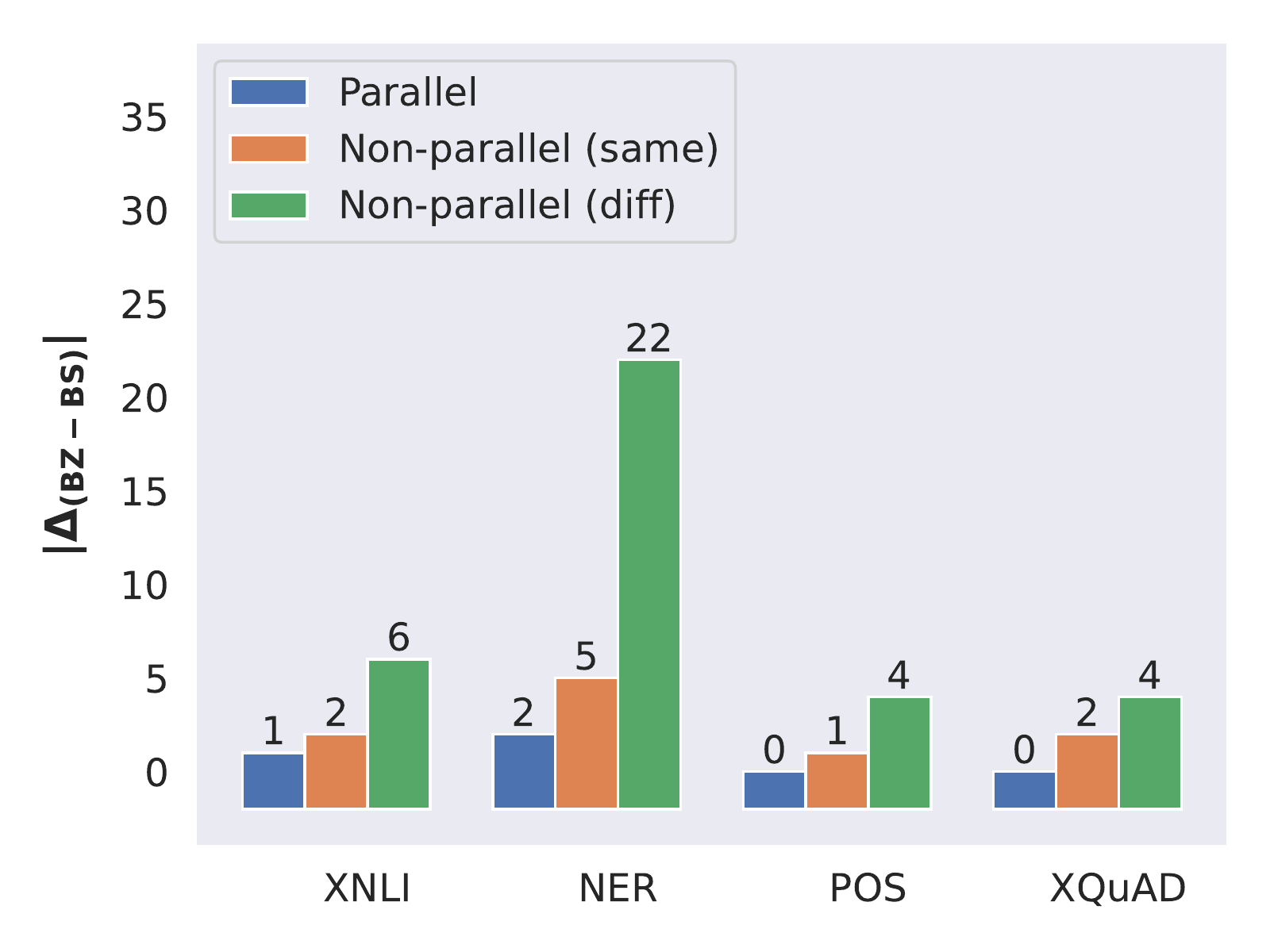}
\caption{
$| \textrm{\supdiff{}} |$ for \script{} under different conditions on the source of \textit{original} and \textit{derived} language pre-training corpora (hereon, corpora)~\symbolsecref{sec:parallel}, averaged over four languages.
Larger values imply worse zero-shot transfer.
The breakdown of scores for different languages is in Appendix~\ref{app:diff_corpus}.
(1) \textit{Non-parallel (diff)} (green bar), which uses corpora from different domains is worse than
(2) \textit{Non-parallel (same)} (orange bar), which uses \underline{different} sets of sentences sampled from the same domain, which is in turn worse than
(3) \textit{Parallel}, which uses the same sentences.
Having pre-training corpora from the same domain like Wikipedia (\textit{Non-parallel (same)}) gives performance boosts between $2$ points for QA to $17$ points for NER when compared to \textit{Non-parallel (diff)}.
}
\label{fig:diff_corpus}
\end{figure}

\begin{figure}[ht]
\centering
\includegraphics[width=\linewidth]{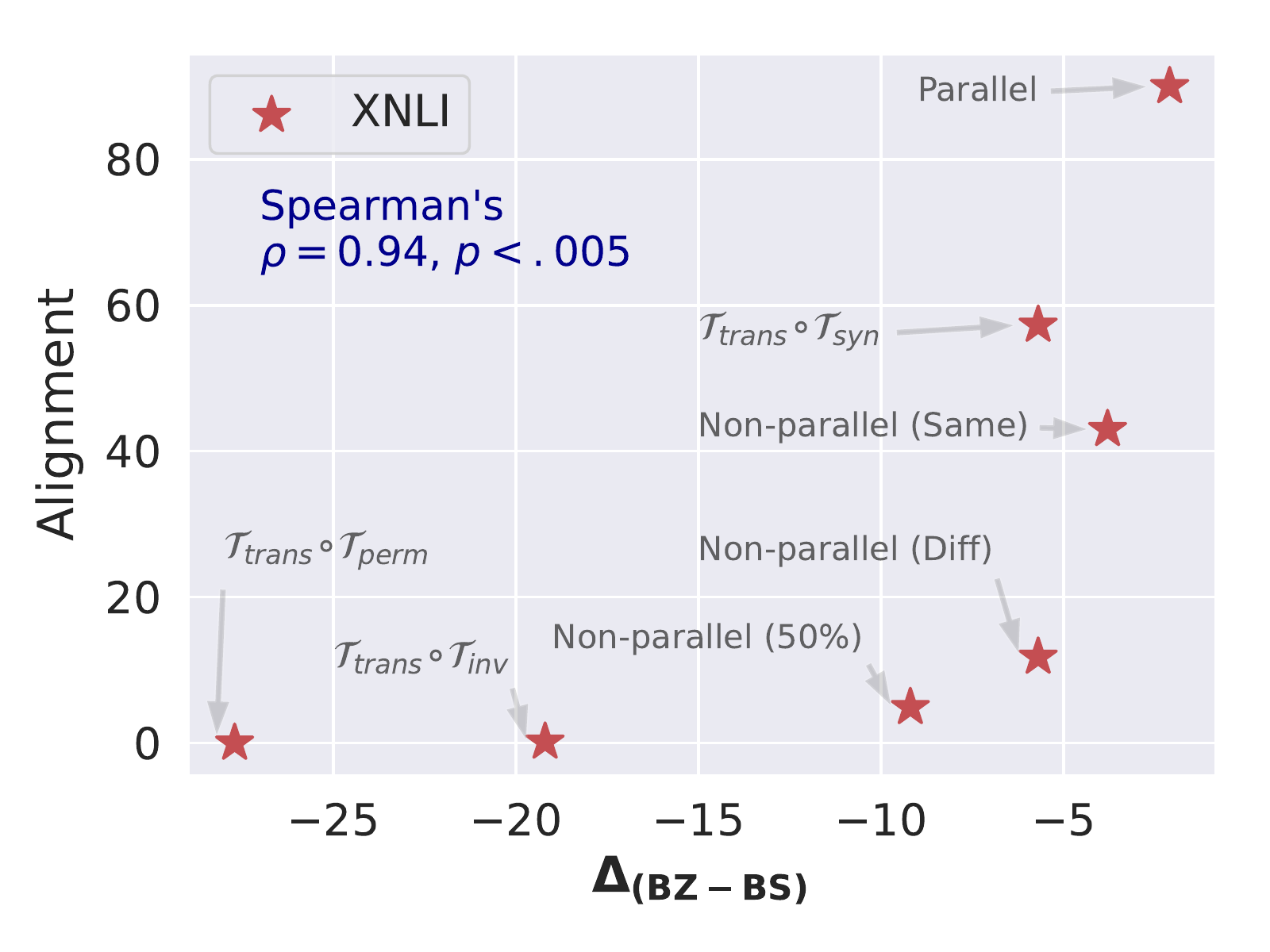}
\caption{
\supdiff{} for \textit{Transliteration} (\script{}) variants on XNLI.
Larger values (less negative) imply better zero-shot transfer.
We see that alignment~\symbolsecref{sec:zs_correlation} between token embeddings of different languages is correlated with~\supdiff{}, and hence with better zero-shot transfer.
For example, \script{} $\circ$ \inv{} (bottom left) which has poor zero-shot transfer also has lower alignment, whereas \textit{Parallel} (top right) which has strong transfer is accompanied with higher alignment.
We find a strong and statistically significant Spearman's correlation of $\rho_s=0.94,p<.005$ on XNLI, $\rho_s=0.93, p<.005$ on NER, and $\rho_s=0.89, p<.01$ on POS.
Plots for other tasks are in Appendix~\ref{app:alignment_correlation}.
}
\label{fig:xnli_correlation_annotated}
\end{figure}

\footnotetext[4]{XQuAD is a question-answering task where the correct answer is a \textit{contiguous} span. We do not report scores on XQuAD for \perm{} and \syntax{} because they can potentially reorder individual words in the contiguous answer, thus distributing them throughout the transformed sentence and making the question unanswerable. On the other hand, \inv{} and \script{} do not have this issue because they maintain the spans.}

\section{Results}
\label{sec:results}

Our experiments reveal several interesting findings for bilingual models including the situational importance of sub-word overlap for zero-shot transfer (\S~\ref{sec:subword1}, \ref{sec:subword2}), 
the effect of domain mismatch between languages (\S~\ref{sec:parallel}),
and correlation of zero-shot performance with embedding alignment (\S~\ref{sec:zs_correlation}).
We connect our findings to zero-shot transfer results between natural languages in Section~\ref{sec:real_languages}.


\subsection{Sub-word overlap is not strictly necessary for strong zero-shot transfer}
\label{sec:subword1}

Sub-word overlap is the number of common tokens between two different language corpora.
If $\mathcal{E}_1$ and $\mathcal{E}_2$ are sets of tokens which appear in the two corpora, then: $\textrm{Sub-word overlap} = \lvert \mathcal{E}_1 \cap \mathcal{E}_2 \rvert / \lvert \mathcal{E}_1 \cup \mathcal{E}_2 \rvert$~\cite{pires2019multilingual}.
The \textit{Transliteration} transformation (\script) creates \textit{original}-\textit{derived} language pairs that have $0\%$ sub-word overlap (equivalently, different scripts), but follow the same word order.

Table~\ref{table:main_table} displays \supdiff{} scores for \script{}, averaged over four languages (Appendix~\ref{appendix:sec:joint_table} contains a breakdown).
We observe that \supdiff{}$\approx 0$ for all tasks while \monodiff{} is highly negative, implying that zero-shot transfer is strong and on par with supervised learning.
This result indicates that zero-shot transfer is possible even when languages with different scripts have similar word orders (in line with~\citet{ketal}).
However, it is unrealistic for natural languages to differ only in their script and not other properties (e.g., word order).


\subsection{Absence of sub-word overlap significantly hurts zero-shot performance when languages differ in their word-orders}
\label{sec:subword2}
To simulate a more realistic scenario, we create \textit{original} and \textit{derived} language pairs which differ both in their scripts ($0\%$ sub-word overlap) and in word order.
We achieve this by composing two transformations on the \textit{original} language corpus, one of which is \textit{Transliteration} (\script).
We experiment with three different compositions, (a) \script{} $\circ$ \inv{}, (b) \script{} $\circ$ \perm{}, and (c) \script{} $\circ$ \syntax{}.
Here, $\boldsymbol{\alpha}$ $\circ$ $\boldsymbol{\beta}$ means that transformation $\boldsymbol{\beta}$ is applied before $\boldsymbol{\alpha}$.
A composed transformation (\script{} $\circ$ $\boldsymbol{\beta}$) differs from its second constituent ($\boldsymbol{\beta}$) in that the former produces a \textit{derived} language which has $0\%$ sub-word overlap with the \textit{original} language whereas the latter has a $100\%$ sub-word overlap.

\paragraph{Results}
Our results (Figure~\ref{fig:composed}, breakdown in Appendix~\ref{app:composed}) show that zero-shot performance is significantly hurt for composed transformations when compared to its constituents.
$|$\supdiff{}$|$ is much larger for \script{} $\circ$ \inv{} when compared to \script{} or \inv{} individually.
For example, for XNLI, $|$\supdiff{}$|$ $=19$ for the composed transformation and just $2$ and $3$ for \script{} and \inv{} individually.
\script{} $\circ$ \perm{} is worse by $\approx 20$ points on XNLI and NER, and over $40$ points on POS when compared to \perm{}.
\script{} $\circ$ \syntax{} suffers lesser than the other two composed transformations, but it is still worse than \syntax{} by $3$, $6$, and $1$ point on XNLI, NER, and POS.
In conclusion, the absence of sub-word overlap significantly degrades zero-shot performance in the realistic case of languages with different word orders.

\subsection{Data from the same domain boosts bilingual performance}
\label{sec:parallel}

Previously, we considered transformations (\trans) that modified the \textit{original} pre-training corpus to get a parallel corpus, $\textrm{\syncorpus} = \textrm{\trans}\left (\textrm{\realcorpus} \right )$, such that there is a one-to-one correspondence between sentences in \realcorpus{} and \syncorpus{} (we call this setting \textit{parallel}).
Since procuring large parallel corpora is expensive in practice, we consider two other settings which use different corpora for \textit{original} and \textit{derived}.

\paragraph{Setup}
Consider two text corpora of the same size, $\mathbf{\mathcal{C}^1_{orig}}$ and $\mathbf{\mathcal{C}^2_{orig}}$.
We compare two settings: (1) The \textit{parallel} setting pre-trains a bilingual model on $\mathbf{\mathcal{C}^1_{orig}} + \mathcal{T} ( \mathbf{\mathcal{C}^1_{orig}}  )$, whereas the (2) \textit{non-parallel} corpus setting uses $\mathbf{\mathcal{C}^1_{orig}} + \mathcal{T} ( \mathbf{\mathcal{C}^2_{orig}}  )$.
We consider two variants of \textit{non-parallel}, (1) \textit{non-parallel (same)} which uses different splits of Wikipedia data (hence, \textit{same} domain), and (2) \textit{non-parallel (diff)} which uses Wikipedia data for the \textit{original} and common crawl data (web text) for the \textit{derived} language (hence, \textit{diff}erent domain).
We use the \textit{Transliteration} transformation (\script{}) to generate the \textit{derived} language corpus and report $|$\supdiff{}$|$ averaged over all languages in Figure~\ref{fig:diff_corpus}.

\paragraph{Results}
We observe consistently on all tasks that the \textit{parallel} setting (blue bar) performs better than both the non-parallel settings.
\textit{Non-parallel (same)} performs better than \textit{non-parallel (diff)}, with gains ranging between $2$ points on XQuAD to $17$ points on NER.
This result shows that even for \textit{original} and \textit{derived} language pairs which differ only in their script, having parallel pre-training corpora leads to the best zero-shot transfer.
Since large-scale parallel unsupervised data is hard to procure, the best alternative is to use corpora from similar domains (Wikipedia) rather than different ones (Wikipedia v.s. web text).

\subsection{Zero-shot performance is strongly correlated with embedding alignment}
\label{sec:zs_correlation}

Our previous results ($\S$~\ref{sec:subword2},~\ref{sec:parallel}) showed cases where zero-shot transfer between languages is poor when there is no sub-word overlap.
To investigate this further, we analyze the static word embeddings learned by bilingual models and find that zero-shot transfer between languages is strongly correlated with the alignment between word embeddings for the \textit{original} and \textit{derived} languages.

\paragraph{Setup}
The \textit{original} and the \textit{derived} languages have a one-to-one correspondence between their sub-word vocabularies when we use \textit{transliteration} (\script).
For a token embedding in the \textit{original}-language embedding matrix, its alignment score is $100\%$ if it retrieves the corresponding token embedding in the \textit{derived} language when a nearest-neighbor search is performed, and $0\%$ otherwise.
We average the alignment score over all the tokens and call it \textit{alignment}.

\paragraph{Results}
We measure the \textit{alignment} of bilingual models pre-trained on different \textit{original}-\textit{derived} language pairs created using \textit{transliteration}, namely the composed transformations~\symbolsecref{sec:subword2}, \textit{parallel}, and \textit{non-parallel}~\symbolsecref{sec:parallel}.
We plot the \textit{alignment} along with the corresponding \supdiff{} scores for XNLI in Figure~\ref{fig:xnli_correlation_annotated}.
Results for other tasks are in Appendix~\ref{app:alignment_correlation}.

We observe that higher \textit{alignment} is associated with lower \supdiff{}, implying better zero-shot transfer.
\textit{Alignment} is lower for composed transformations like \script{} $\circ$ \inv{} and \script{} $\circ$ \perm{} which have large and negative \supdiff{}.
\textit{Alignment} also explains the results in Section~\ref{sec:parallel}, with \textit{non-parallel} variants having lower alignment scores than \textit{parallel}, which is in line with their lower \supdiff{}.
Overall, we find a strong and significant Spearman's rank correlation between \textit{alignment} and \supdiff{}, with $\rho=0.94, p<.005$ for XNLI, $\rho=0.93, p<.005$ for NER, and $\rho=0.89, p<.01$ for POS, indicating that increasing the embedding alignment between languages helps improve zero-shot transfer.

\subsection{Connections to results on natural language pairs}
\label{sec:real_languages}

\paragraph{Effect of sub-word overlap}
In \S~\ref{sec:subword2}, we showed that when languages have different scripts ($0\%$ sub-word overlap), zero-shot transfer significantly degrades when they additionally have different word orders.
However, the zero-shot transfer is good when languages differ only in the script and have similar or the same word order.
This is in line with anecdotal evidence in~\citet{pires2019multilingual}, where zero-shot transfer works well between \textit{English} and \textit{Bulgarian} (different script but same subject-verb-object order -- SVO), but is poor between \textit{English} and \textit{Japanese} (different script \textit{and} word order -- SVO v.s. SOV).
Our result also corroborates findings in \citet{conneau2020emerging} that artificially increasing sub-word overlap between natural languages (which have different word orders) improves performance (e.g., $3$ points on XNLI).

\paragraph{Effect of token embedding alignment}
In \S~\ref{sec:zs_correlation}, we showed that zero-shot transfer is strongly correlated with word embedding alignment between languages.
This explains the usefulness of recent studies which try to improve multilingual pre-training with the help of auxiliary objectives, which improve word or sentence embedding alignment.

DICT-MLM~\cite{DBLP:journals/corr/abs-2010-12566} and RelateLM~\cite{khemchandani-etal-2021-exploiting} require the model to predict cross-lingual synonyms as an auxiliary objective, thus indirectly improving word-embedding alignment and the zero-shot performance on multiple tasks.
\citet{hu2021explicit} add an auxiliary objective that implicitly improves word embedding alignment and show that they can achieve performance similar to larger models.
\citet{cao2019multilingual} explicitly improve contextual word embedding alignment with the help of word-level alignment information in machine-translated cross-lingual sentence pairs.
Since they apply this post hoc and not during pre-training, the improvement, albeit significant, is small ($2$ points on XNLI).
While these studies do not fully utilize word and sentence embedding alignment information, our results lead us to posit that they are a step in the right direction and that baking alignment information more explicitly into pre-training will be beneficial.

\section{Conclusion}
\label{sec:conclusion}


Through a systematic study of zero-shot transfer between four diverse natural languages and their counterparts created by modifying specific properties like the script, word order, and syntax, we showed that (1) absence of sub-word overlap hurts zero-shot performance when languages differ in their word order, and (2) zero-shot performance is strongly correlated with word embedding alignment between languages.
Some recent studies have implicitly or unknowingly attempted to improve alignment and have shown slight improvements in zero-shot transfer performance.
However, our results lead us to posit that explicitly improving word embedding alignment during pre-training by using either supervised (e.g., parallel sentences and translation dictionaries) or unsupervised data will significantly improve zero-shot transfer.
Although real languages typically differ in more ways than the set of properties considered in our transformations, our methodology is still useful to help isolate crucial properties for transfer. Future work can experiment with more sophisticated transformations and investigate closer connections with human language pairs.

\section*{Acknowledgments}
This work was funded through a grant from the Chadha Center for Global India at Princeton University. We thank Shunyu Yao, Vishvak Murahari, Tianyu Gao, Sadhika Malladi, and Jens Tuyls for reading our draft and providing valuable comments.
We also thank the Google Cloud Research program for computational support in running our experiments.

\bibliography{custom}
\bibliographystyle{acl_natbib}

\clearpage
\newpage
\appendix
\section*{Appendices}

\section{Mathematical Specification for Transformation of Downstream Datasets} \label{appendix:MathematicalDownstream}

\paragraph{Text classification}

Text classification tasks like news classification or sentiment analysis typically have instances which contain a single sentence and a label.
Instances in other classification tasks like natural language inference (NLI)~\cite{bowman2015recursive} contain two sentences and one label.
For such tasks, we apply the transformation (\trans) on each sentence within every instance, and leave the annotated label as is.
Therefore, for a dataset of size $n$ which contains $m$ sentences per instance, we have:
\begin{align*}
    \begin{split}
        \textrm{\realdown{}} &= \{ \left ( s_{i1}, \dots, s_{im}, y_i \right ) \; \vert \; i=1:N \} \\
        \textrm{\syndown{}} &= \{ \left ( \mathcal{T}(s_{i1}), \dots, \mathcal{T}(s_{im}), y_i \right ) \; \vert \; i=1:N \} \\
    \end{split}
\end{align*}

\paragraph{Token-classification tasks}
Tasks like named-entity recognition (NER) and part-of-speech tagging (POS tagging) have labels associated with \textit{each} token in the sentence.
For these datasets, we ensure that any transformation (\trans) that changes the order of the tokens also changes the order of the corresponding labels.

We define a few quantities to express the transformation mathematically.
Let $s_i = \left ( w_{i1}, \dots, w_{ik} \right )$ be a sentence comprised of $k$ tokens and $y_i = \left ( y_{i1}, \dots, y_{ik} \right )$ be labels corresponding to the tokens in the sentence.
We define a new transformation ($\mathcal{T}_{aug}$) which operates on the label augmented sentence, $s_i^{aug} = \left ( \left ( w_{i1}, y_{i1} \right ) , \dots, \left ( w_{ik}, y_{ik} \right ) \right )$.
Let $s_i^{aug}[j]$ correspond to the $j^{th}$ element in the sequence, and $s_i^{aug}[j][\texttt{word}]$ and $s_i^{aug}[j][\texttt{label}]$ correspond to the word and label of the $j^{th}$ element.
Let $\mathcal{T}_{aug}(s_i^{aug})[j][\texttt{orig}]$ denote the index of the $j^{th}$ element in the transformed sequence with respect to the original sequence $s_i^{aug}$.
Then, the new transformation $\mathcal{T}_{aug}$ is such that,
\begin{align*}
    \begin{split}
        &\mathcal{T}_{aug}(s_i^{aug})[j][\texttt{orig}] = \mathcal{T}(s_i)[j][\texttt{orig}] \\
        &\textrm{Let}\;\; \texttt{orig\_j} = \mathcal{T}_{aug}(s_i^{aug})[j][\texttt{orig}] \\
        &\mathcal{T}_{aug}(s_i^{aug})[j][\texttt{label}] = s_i^{aug}[\texttt{orig\_j}][\texttt{label}]
    \end{split}
\end{align*}
We transform the dataset using $\mathcal{T}_{aug}$:
\begin{align*}
    \begin{split}
        \textrm{\realdown{}} &= \{ s_i^{aug} \; \vert \; i=1:N \} \\
        \textrm{\syndown{}} &= \{ \mathcal{T}_{aug}(s_i^{aug}) \; \vert \; i=1:N \} \\
    \end{split}
\end{align*}

\section{Zero-shot transfer results for different transformations}
\label{appendix:sec:joint_table}

\begin{table*}[t]
\centering
\resizebox{\linewidth}{!}{
\begin{tabular}{cccccccccccccc}\toprule
\multirow{2}{*}{\textbf{Task}} & \multirow{2}{*}{\textbf{Language}} & \multicolumn{3}{c}{\textbf{Inversion}} & \multicolumn{3}{c}{\textbf{Permutation}} & \multicolumn{3}{c}{\textbf{Syntax}} & \multicolumn{3}{c}{\textbf{Transliteration}}\\
\cmidrule(lr){3-5} \cmidrule(lr){6-8} \cmidrule(lr){9-11} \cmidrule(lr){12-14} 

 & & \zs & \supdiff & \monodiff & \zs & \supdiff & \monodiff & \zs & \supdiff & \monodiff & \zs & \supdiff & \monodiff\\ \midrule


\multirow{4}{*}{\textbf{XNLI}} & \textbf{English} & 73.2 & \gradient{-3.4} & \gradient{-14.9} & 68.6 & \gradient{-5} & \gradient{-7.7} & 74.1 & \gradient{-1.8} & \gradient{-1.5} & 74.1 & \gradient{-1.7} & \gradient{-42.5}\\
& \textbf{French} & 62.5 & \gradient{-9.5} & \gradient{-8.8} & 68.4 & \gradient{-1} & \gradient{-7.6} & 69.6 & \gradient{-2.2} & \gradient{-1.4} & 71.6 & \gradient{-1.6} & \gradient{-39.9}\\
& \textbf{Hindi} & 43.9 & \gradient{-15.7} & \gradient{-15.8} & 51.2 & \gradient{-6.2} & \gradient{-13.1} & 61.6 & \gradient{-0.3} & \gradient{-1.6} & 63.4 & \gradient{-0.1} & \gradient{-29.4}\\
& \textbf{Arabic} & 54 & \gradient{-12.3} & \gradient{-12.5} & 62.1 & \gradient{-2.3} & \gradient{-6} & 65.9 & \gradient{0.7} & \gradient{0.3} & 68 & \gradient{-0.4} & \gradient{-35.1} \\ \cdashlinelr{1-14}
& \textbf{Avg.} & 58.4  & \textbf{\gradient{-10.2}} & \textbf{\gradient{-13}} & 62.6 & \textbf{\gradient{-3.6}} & \textbf{\gradient{-8.6}} & 67.8 & \textbf{\gradient{-0.9}} & \textbf{\gradient{-1.1}} & 69.3 & \textbf{\gradient{-1.0}} & \textbf{\gradient{-36.7}} \\ \midrule
 
\multirow{4}{*}{\textbf{NER}} & \textbf{English} & 39.8 & \gradient{-44.5} & \gradient{-35.9} & 40.2 & \gradient{-28.5} & \gradient{-33.2} & 61.1 & \gradient{-7.8} & \gradient{-10.3} & 78 & \gradient{-2.1} & \gradient{-70.2} \\
& \textbf{French} & 54.5 & \gradient{-34.4} & \gradient{-51.3} & 44.4 & \gradient{-36.0} & \gradient{-39.8} & 59.6 & \gradient{-21.9} & \gradient{-25.9} & 84.3 & \gradient{-3.1} & \gradient{-87.4} \\
& \textbf{Hindi} & 19.4 & \gradient{-63.9} & \gradient{-63.2} & 38.5 & \gradient{-21.9} & \gradient{-37.4} & 64.8 & \gradient{-8.4} & \gradient{-7.3} & 84.4 & \gradient{-0.5} & \gradient{-82.9} \\
& \textbf{Arabic} & 37.8 & \gradient{-53.6} & \gradient{-36.3} & 66.2 & \gradient{-18.8} & \gradient{-31.1} & 66.1 & \gradient{-20.1} & \gradient{-23} & 88 & \gradient{-1.9} & \gradient{-89.9} \\  \cdashlinelr{1-14}
& \textbf{Avg.} & 37.9  & \textbf{\gradient{-49.1}} & \textbf{\gradient{-46.7}} & 47.3 & \textbf{\gradient{-26.3}} & \textbf{\gradient{-35.4}} & 62.9 & \textbf{\gradient{-14.6}} & \textbf{\gradient{-16.6}} & 83.7 & \textbf{\gradient{-1.9}} & \textbf{\gradient{-82.6}} \\ \midrule
 
\multirow{4}{*}{\textbf{POS}} & \textbf{English} & 94.4 & \gradient{-0.7} & \gradient{-24.3} & 78.3 & \gradient{-11.9} & \gradient{-17.6} & 92.9 & \gradient{-0.9} & \gradient{-2.2} & 94.6 & \gradient{-0.5} & \gradient{-95.1}\\
& \textbf{French} & 74.3 & \gradient{-22.7} & \gradient{-22.9} & 82 & \gradient{-12.2} & \gradient{-20.9} & 93.5 & \gradient{-3.2} & \gradient{-5.2} & 97.2 & \gradient{-0.2} & \gradient{-97.4}\\
& \textbf{Hindi} & 19 & \gradient{-74.5} & \gradient{-74.5} & 51 & \gradient{-14} & \gradient{-41.8} & 91.6 & \gradient{-3.3} & \gradient{-11.3} & 96.5 & \gradient{-0.1} & \gradient{-96.6} \\
& \textbf{Arabic} & 69.2 & \gradient{-23} & \gradient{-23} & 83.1 & \gradient{-6.5} & \gradient{-20.6} & 79.4 & \gradient{-10} & \gradient{-11.5} & 93.2 & \gradient{-0.8} & \gradient{-90.9} \\ \cdashlinelr{1-14}
& \textbf{Avg.} & 64.2 & \textbf{\gradient{-30.2}} & \textbf{\gradient{-36.2}} & 73.6 & \textbf{\gradient{-11.2}} & \textbf{\gradient{-25.2}} & 89.4 & \textbf{\gradient{-4.4}} & \textbf{\gradient{-7.6}} & 95.4  & \textbf{\gradient{-0.4}} & \textbf{\gradient{-95.0}} \\ \midrule

\multirow{4}{*}{\textbf{XQuAD}} & \textbf{English} & 30.4 & \gradient{-43.2} & \gradient{-35.5} & - & - & - & - & - & - & 72.4 & \gradient{-4} & \gradient{-73} \\
& \textbf{French} & 25.2 & \gradient{-29.5} & \gradient{-29.6} & - & - & - & - & - & - & 60.9 & \gradient{-1} & \gradient{-55.5} \\
& \textbf{Hindi} & 14.5 & \gradient{-27.3} & \gradient{-27.3} & - & - & - & - & - & - & 57.3 & \gradient{10.6} & \gradient{-43.5} \\
& \textbf{Arabic} & 21 & \gradient{-31.2} & \gradient{-31.4} & - & - & - & - & - & - & 54 & \gradient{-0.5} & \gradient{-51.7} \\ \cdashlinelr{1-14}
& \textbf{Avg.} & 22.8 & \textbf{\gradient{-32.8}} & \textbf{\gradient{-31.0}} &  &  &  &  &  &  & 61.2 & \textbf{\gradient{1.3}} & \textbf{\gradient{-55.9}}\\

\bottomrule
\end{tabular}
}
\caption{
\label{table:joint_table}
This table is an extended version of Table~\ref{table:main_table} in the main paper.
Smaller (more negative) \supdiff{} implies worse bilingual zero-shot transfer, whereas \supdiff{}$\approx 0$ implies strong transfer.
\supdiff{} $\gg$ \monodiff{} implies that bilingual pre-training is extremely useful.
Scores are highlighted based on their value (lower scores have a higher intensity of {\setlength{\fboxsep}{2pt}\colorbox{low!40}{red}}).
\textbf{(1) Discussing \supdiff{}:}
\script{} exhibits strong transfer on all tasks and languages (high \supdiff{} scores), and bilingual pre-training is extremely useful (\supdiff{} $\gg$ \monodiff{}), implying that zero-shot transfer is possible between languages with different scripts but the same word order.
\inv{} and \perm{} suffer on all tasks (small \supdiff{} scores) whereas \syntax{} suffers significantly lesser, which provides evidence that local changes to the word order made by \textit{Syntax} (\syntax{}) hurts zero-shot transfer significantly lesser than global changes made by \textit{Inversion} (\inv{}) and \textit{Permutation} (\perm{}).
\textbf{(1) Discussing \monodiff{}:}
\supdiff{} is much larger than \monodiff{} for \script{}, implying that bilingual pre-training (hereon, pre-training) is extremely useful.
\supdiff{} and \monodiff{} are similar for \inv{} and \syntax{}, implying that pre-training is not beneficial for these transformations.
\supdiff{} is slightly larger than \monodiff{} for \perm{}, which means that pre-training is moderately useful.
}
\end{table*}

Table~\ref{table:joint_table} in the appendix is the extended version of Table~\ref{table:main_table} in the main paper with a breakdown for all languages.
It reports \supdiff{}, \monodiff{}, and \zs{} for different languages and transformations considered.

\section{Composed Transformations}
\label{app:composed}

Table~\ref{table:Composition} in the appendix presents the breakdown of results in Figure~\ref{fig:composed} of the main paper.
It reports \supdiff{} scores for composed transformations and their constituents.

\begin{table}[t]
\centering
\resizebox{\linewidth}{!}{
\begin{tabular}{ccccccc}\toprule

 \multirow{2}{*}{\textbf{\trans}} & \multicolumn{2}{c}{\textbf{XNLI}} & \multicolumn{2}{c}{\textbf{NER}} & \multicolumn{2}{c}{\textbf{POS}} \\
\cmidrule(lr){2-3} \cmidrule(lr){4-5} \cmidrule(lr){6-7}
  & \graytext{\zs} & \supdiff & \graytext{\zs} & \supdiff & \graytext{\zs} & \supdiff \\ \midrule
  \script & \graytext{74.1} & -2.1 & \graytext{78} & -2.3 & \graytext{94.6} & -0.5 \\ \cdashlinelr{1-7}
  \inv & \graytext{73.2} & -3.4 & \graytext{39.8} & -44.5 & \graytext{94.4} & -0.7 \\
 \script{} $\circ$ \inv{} & \graytext{55.7} & -19.2 & \graytext{32.5} & -51.5 & \graytext{52.2} & -42.7 \\ \cdashlinelr{1-7}
  \perm & \graytext{68.6} & -5 & \graytext{40.2} & -28.5 & \graytext{78.3} & -11.9 \\
 \script{} $\circ$ \perm{} & \graytext{44} & -27.7 & \graytext{17.1} & -46.3 & \graytext{29.5} & -59 \\ \cdashlinelr{1-7}
 \syntax & \graytext{74.1} & -1.8 & \graytext{61.1} & -7.8 & \graytext{92.9} & -0.9 \\
 \script{} $\circ$ \syntax{} & \graytext{69.8} & -5.7 & \graytext{53.5} & -14.2 & \graytext{91.5} & -2 \\  
\bottomrule
\end{tabular}
}
\caption{
\label{table:Composition}
Breakdown of results in Figure~\ref{fig:composed} of the main paper.
\zs{} is the zero-shot performance.
\supdiff{}, \monodiff{}, and \zs{} are described in Section~\ref{subsection:Evaluation} and Table~\ref{table:EvalExplanation}.
Composing transformations always hurts \supdiff{} when compared to individual transformations.
}
\end{table}

\section{Comparing different sources for \textit{original} and \textit{derived} language corpora}
\label{app:diff_corpus}

\begin{table}[t]
\centering
\resizebox{\linewidth}{!}{
\begin{tabular}{ccccc}\toprule

\textbf{Transliteration} & \multicolumn{3}{c}{\textbf{\supdiff}\; $(\uparrow)$} & \multirow{2}{*}{\textbf{Alignment $(\uparrow)$}} \\ \cmidrule(lr){2-4}
\textbf{Variant} & \textbf{XNLI} & \textbf{NER} & \textbf{POS} &  \\ \midrule
\textbf{Parallel} & -2.1 & -2.3 & -0.5 & 90.0 \\ \cdashlinelr{1-5}
\textbf{Trans \comp{} Syntax} & -5.7 & -14.2 & -2 & 57.3 \\ \cdashlinelr{1-5}
\textbf{Non-parallel} & \multirow{2}{*}{-3.8} & \multirow{2}{*}{-4.1} & \multirow{2}{*}{-0.7} & \multirow{2}{*}{43.0} \\
\textbf{(Same)} & & & & \\ \cdashlinelr{1-5}
\textbf{Non-parallel} & \multirow{2}{*}{-5.7} & \multirow{2}{*}{-14.3} & \multirow{2}{*}{-1.5} & \multirow{2}{*}{11.8} \\
\textbf{(Diff)} & & & & \\ \cdashlinelr{1-5}
\textbf{Trans \comp{} Inv} & -19.2 & -51.5 & -42.7 & 0.16 \\ \cdashlinelr{1-5}
\textbf{Trans \comp{} Perm} & -27.7 & -46.3 & -59 & 0.01 \\ 

\bottomrule
\end{tabular}
}
\caption{
\supdiff{} and \textit{alignment} scores for different \textit{Transliteration} variants.
The table contains raw scores for results in Section~\ref{sec:zs_correlation} of the main paper.
Rows are sorted in descending order based on \textit{alignment}.
We observe strong correlations between alignment and zero-shot transfer, with $\rho_s=0.94,p<.005$ on XNLI, $\rho_s=0.93, p<.005$ on NER, and $\rho_s=0.89, p<.01$ on POS.
}
\label{table:Alignment}
\end{table}

\begin{table*}[t]
\centering
\begin{tabular}{cccccc}\toprule
\multirow{2}{*}{\textbf{Task}} & \multirow{2}{*}{\textbf{Language}} & \textbf{XNLI} & \textbf{NER} & \textbf{POS} & \textbf{XQuAD}\\
\cmidrule(lr){3-3} \cmidrule(lr){4-4} \cmidrule(lr){5-5} \cmidrule(lr){6-6} 

 & & \supdiff & \supdiff & \supdiff & \supdiff\\ \midrule


\multirow{4}{*}{\textbf{Parallel}} & \textbf{English} & -1.7 & -2.1 & -0.5 & -4 \\
& \textbf{French} & -1.6 & -3.1 & -0.2 & -1 \\
& \textbf{Hindi} & -0.1 & -0.5 & -0.1 & 10.6 \\
& \textbf{Arabic} & -0.4 & -1.9 & -0.8 & -0.5  \\ \cdashlinelr{1-6}
& \textbf{Avg.} & \textbf{-1.0} & \textbf{-1.9} & \textbf{-0.4} & \textbf{1.3} \\ \midrule

\multirow{4}{*}{\textbf{Non-parallel (Same)}} & \textbf{English} & -3.8 & -4.1 & -0.7 & -6.9 \\
& \textbf{French} & -1 & -6.3 & -0.5 & -0.9 \\
& \textbf{Hindi} & -0.4 & -3.1 & -0.2 & 4.5 \\
& \textbf{Arabic} & -2 & -6.1 & -1.5 & 0.7  \\ \cdashlinelr{1-6}
& \textbf{Avg.} & \textbf{-1.8} & \textbf{-4.9} & \textbf{-0.7} & \textbf{-0.6}  \\ \midrule

\multirow{4}{*}{\textbf{Non-parallel (Diff)}} & \textbf{English} & -5.7 & -14.3 & -1.5 & -9.3 \\
& \textbf{French} & -10.9 & -30.3 & -10.5 & -5.2 \\
& \textbf{Hindi} & -0.5 & -8.6 & -1 & 5 \\
& \textbf{Arabic} & -6.3 & -34.7 & -3.7 & -1.9  \\ \cdashlinelr{1-6}
& \textbf{Avg.} & \textbf{-5.9} & \textbf{-22.0} & \textbf{-4.2} & \textbf{-2.9} \\

\bottomrule
\end{tabular}
\caption{
\label{app:table:diff_corpus_extended}
$| \textrm{\supdiff{}} |$ for \script{} under different conditions on the source of \textit{original} and \textit{derived} language pre-training corpora ~\symbolsecref{sec:parallel}.
Larger values imply worse zero-shot transfer.
For all languages:
(1) \textit{Non-parallel (diff)}, which uses corpora from different domains is worse than
(2) \textit{Non-parallel (same)}, which uses \underline{different} sets of sentences sampled from the same domain, which is in turn worse than
(3) \textit{Parallel}, which uses the same sentences.
}
\end{table*}

Table~\ref{app:table:diff_corpus_extended} in the appendix contains the breakdown of results in Figure~\ref{fig:diff_corpus} of the main paper.
It reports \supdiff{} for different languages on different tasks for the settings mentioned in Section~\ref{sec:parallel}.

\section{Alignment Correlation}
\label{app:alignment_correlation}

\begin{figure*}[ht]

\centering
\begin{subfigure}[b]{0.45\textwidth}
\centering
\includegraphics[width=\textwidth]{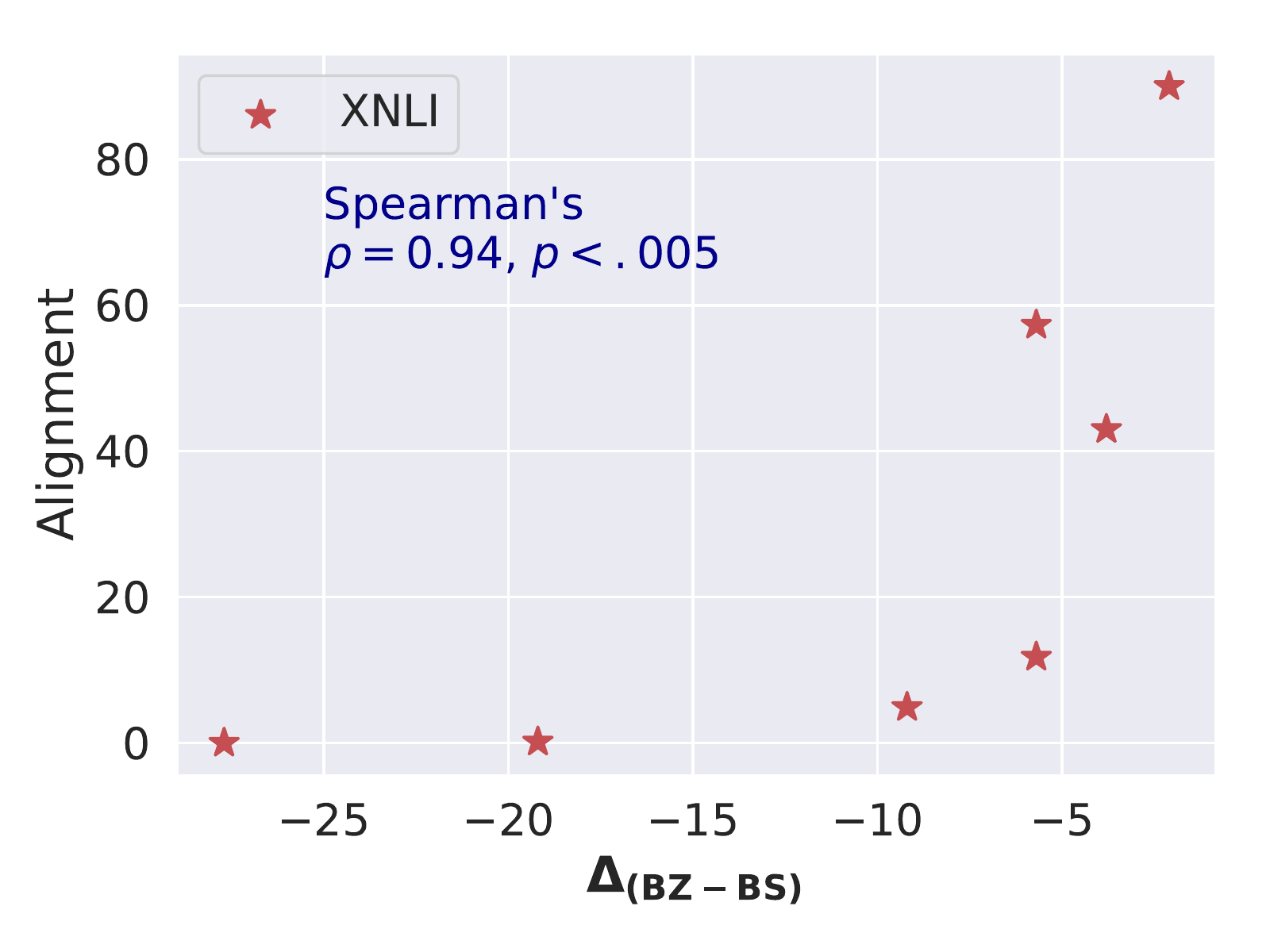}
\end{subfigure}
\begin{subfigure}[b]{0.45\textwidth}
\centering
\includegraphics[width=\textwidth]{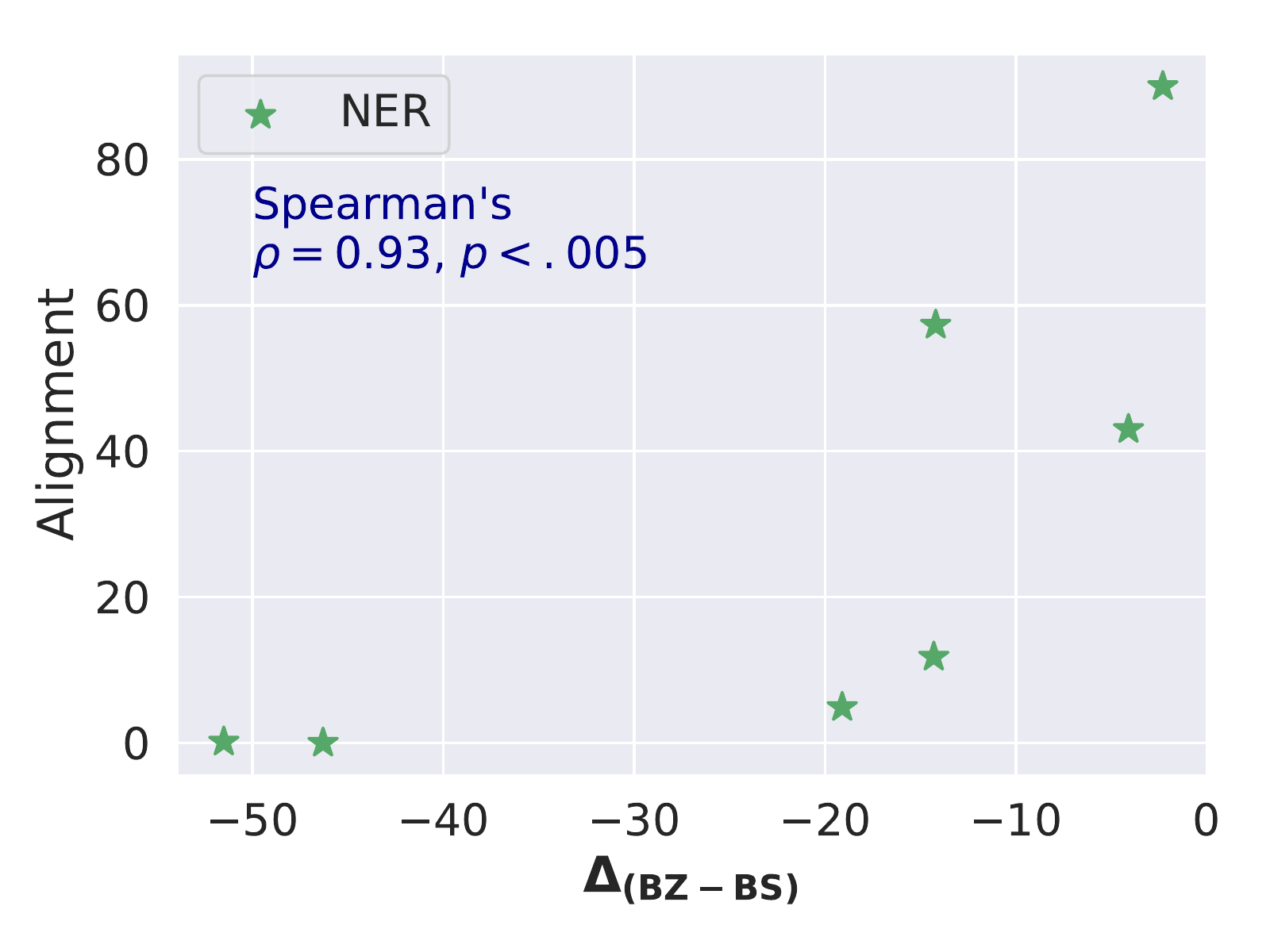}
\end{subfigure}
\begin{subfigure}[b]{0.45\textwidth}
\centering
\includegraphics[width=\textwidth]{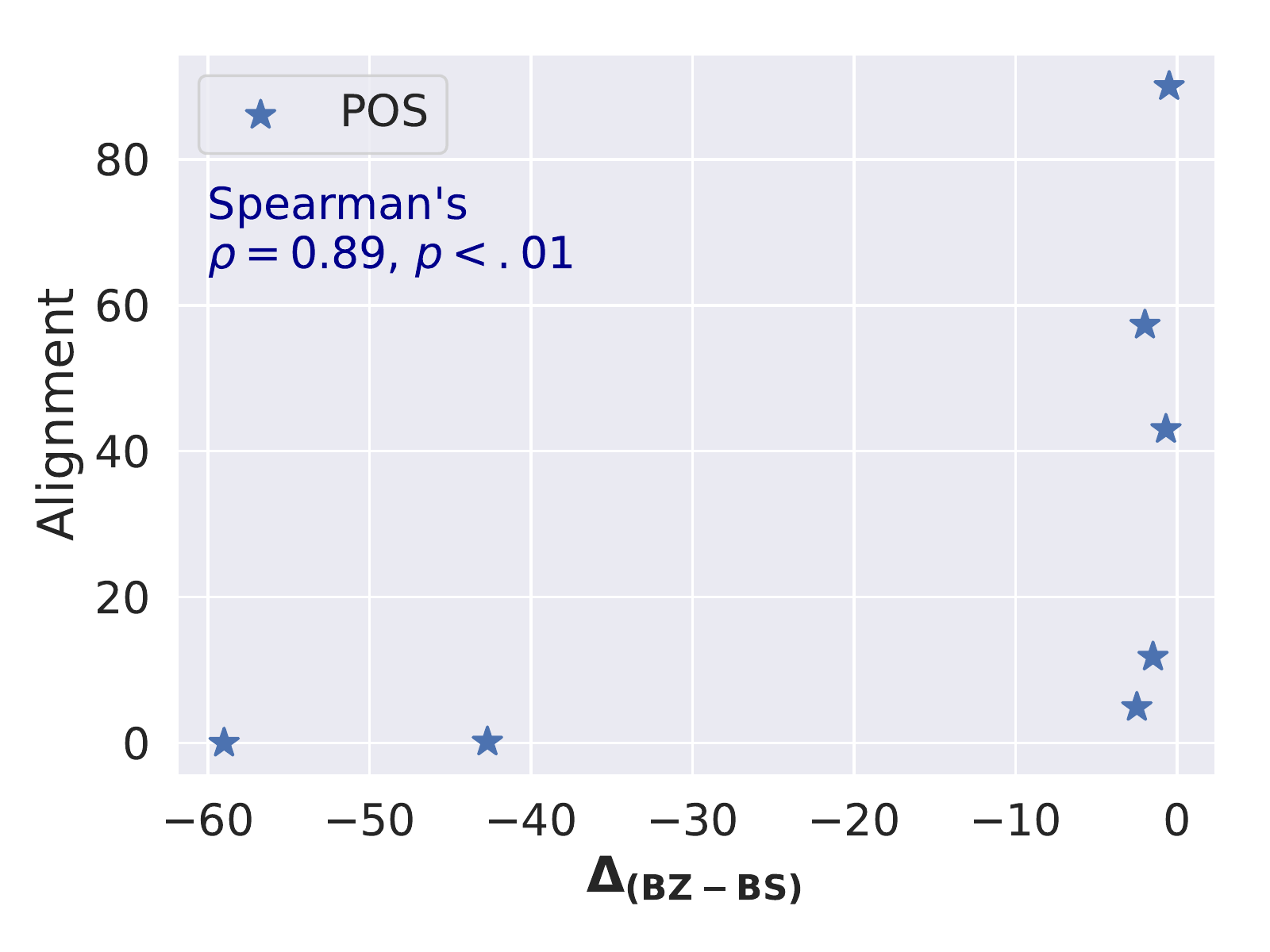}
\end{subfigure}

\caption{
Alignment v.s. \supdiff{} plots for XNLI, NER, and POS.
We observe strong correlations between alignment and zero-shot transfer, with $\rho_s=0.94,p<.005$ on XNLI, $\rho_s=0.93, p<.005$ on NER, and $\rho_s=0.89, p<.01$ on POS.
}
\label{app:fig:composed}

\end{figure*}

We present alignment results (Section~\ref{sec:zs_correlation}) for all XNLI, NER, and POS in Figure~\ref{app:fig:composed}.
We observe strong correlations between alignment and zero-shot transfer, with $\rho_s=0.94,p<.005$ on XNLI, $\rho_s=0.93, p<.005$ on NER, and $\rho_s=0.89, p<.01$ on POS.
We present the raw scores in Table~\ref{table:Alignment}.

\section{Hyperparameters for XTREME}
\label{appendix:xtreme}

\begin{itemize}
    \item XNLI: Learning rate -- \texttt{2e-5}, maximum sequence length -- 128, epochs -- 5, batch size -- 32.
    \item NER: Learning rate -- \texttt{2e-5}, maximum sequence length -- 128, epochs -- 10, batch size -- 32.
    \item POS: Learning rate -- \texttt{2e-5}, maximum sequence length -- 128, epochs -- 10, batch size -- 32.
    \item Tatoeba: Maximum sequence length -- 128, pooling strategy -- representations from the middle layer $\left ( \frac{n}{2} \right )$ of the model.
    \item XQuAD: Learning rate -- \texttt{3e-5}, maximum sequence length -- 384, epochs -- 2, document stride -- 128, warmup steps -- 500, batch size -- 16, weight decay -- 0.0001.
\end{itemize}

\end{document}